\documentclass[10pt]{article}

\usepackage[letterpaper,top=0.55in,bottom=0.68in,left=1.50in,right=1.50in]{geometry}
\newcommand{\method}{StreamingWAM}

\usepackage[T1]{fontenc}
\usepackage[utf8]{inputenc}
\usepackage{times}
\usepackage[round,authoryear]{natbib}

\usepackage{amsmath}
\usepackage{amssymb}
\usepackage{amsthm}
\usepackage{mathtools}

\usepackage{graphicx}
\usepackage{booktabs}
\usepackage{multirow}
\usepackage{subcaption}
\usepackage{wrapfig}
\usepackage{float}
\usepackage{xcolor}

\usepackage{algorithm}
\usepackage{algpseudocode}

\usepackage{microtype}
\usepackage{url}
\usepackage{xspace}

\usepackage{hyperref}
\hypersetup{
  colorlinks=true,
  linkcolor=blue!60!black,
  citecolor=blue!60!black,
  urlcolor=blue!60!black,
  pdftitle={StreamingWAM: Streaming World Action Modeling via Chunk-Wise Staircase Flow Matching},
  pdfauthor={Shaojie Ren, Jingwen Sun, Zezhi Liu, Hongjin Chen, Xin Jin, Wei Zou, and Zhibo Chen}
}
\usepackage[capitalize,noabbrev]{cleveref}

\newcommand{\fastwam}{\mbox{Fast-WAM}\xspace}
\newcommand{\fwjoint}{\mbox{Fast-WAM-Joint}\xspace}
\newcommand{\fasterwam}{\mbox{Faster-WAM}\xspace}
\newcommand{\liberoplus}{\textsc{LIBERO-Plus}\xspace}
\newcommand{\robotwin}{\textsc{RoboTwin~2.0}\xspace}

\newcommand{\Kexec}{K}

\newcommand{\nprefix}{P}
\newcommand{\obs}{o}
\newcommand{\act}{a}

\newcommand{\parhead}[1]{\paragraph{#1}}

\title{StreamingWAM: Streaming World Action Modeling via Chunk-Wise Staircase Flow Matching}
\author{Shaojie Ren, Jingwen Sun, Zezhi Liu, Hongjin Chen, Xin Jin, Wei Zou, and Zhibo Chen}
\date{}

\begin{document}
\raggedbottom

\begin{center}
{\LARGE\bfseries StreamingWAM: Streaming World Action Modeling\\
via Chunk-Wise Staircase Flow Matching\par}
\vspace{0.8em}
{\normalsize
Shaojie Ren\textsuperscript{1,2,3,*}\quad
Jingwen Sun\textsuperscript{4,*}\quad
Zezhi Liu\textsuperscript{5}\quad
Hongjin Chen\textsuperscript{6}\\[-0.05em]
Xin Jin\textsuperscript{7}\quad
Wei Zou\textsuperscript{3,\ensuremath{\dagger}}\quad
Zhibo Chen\textsuperscript{4,\ensuremath{\dagger}}\par}
\vspace{0.6em}
{\small
\textsuperscript{1}University of Chinese Academy of Sciences\quad
\textsuperscript{2}Beijing Zhongguancun Academy\\[-0.08em]
\textsuperscript{3}Institute of Automation, Chinese Academy of Sciences\\[-0.08em]
\textsuperscript{4}University of Science and Technology of China\quad
\textsuperscript{5}Nankai University\\[-0.08em]
\textsuperscript{6}Harbin Institute of Technology\quad
\textsuperscript{7}Eastern Institute of Technology, Ningbo\\[0.25em]
\textsuperscript{*}Equal contribution.\quad
\textsuperscript{\ensuremath{\dagger}}Corresponding authors.\\[0.25em]
\href{https://renshaojie233.github.io/StreamingWAM/}{Project page: renshaojie233.github.io/StreamingWAM}
\par}
\end{center}
\vspace{0.25em}
\begin{abstract}
World Action Models (WAMs), often built upon pretrained video-generation world
models, have emerged as a promising framework for general-purpose and
open-ended robot control.  However, their real-world deployment often suffers
from discontinuous and jerky execution.  Existing methods typically address
this issue by either improving inference efficiency or enhancing action
continuity across prediction chunks, without jointly addressing both
challenges.  To bridge this gap, we propose \method{}, a streaming-diffusion
framework for WAM-based robot control.  \method{} combines a noise ladder for
chunk-wise action prediction, a prefix rollout module for cross-chunk
continuity, and dense video-history memory for temporally consistent
conditioning.  Experiments on LIBERO, LIBERO-Plus, RoboTwin~2.0,
and real-world robotic tasks demonstrate that \method{} improves action
continuity across static and dynamic settings, achieving competitive
performance while reducing steady-state replanning latency to 72.9\,ms in
simulation and 57.3\,ms on the real robot.  The method also produces smoother
action transitions, while the streaming staircase reduces sensitivity to the
execution horizon~\(K\).

\end{abstract}
\vspace{1.0em}

\section{Introduction}
\label{sec:intro}

\noindent
Recent WAMs are built upon world models, particularly pretrained video-generation models~\citep{ye2026dreamzero, li2026lingbotva, yuan2026fastwam, kim2026cosmospolicy}. Through pretraining objectives based on future-state prediction, these models acquire knowledge of environment dynamics, which is then adapted to robotic domains via fine-tuning for general-purpose and open-ended robot control~\citep{zhang2026dowambetterthanvlas,wang2026wamsurvey}. By leveraging this learned dynamics prior, WAMs can exhibit strong generalization when deployed on real-world robots across diverse tasks and environments~\citep{wang2026openwam}.

Despite their impressive performance, WAMs often exhibit temporally discontinuous and jerky motions during closed-loop execution~\citep{motus2026rtwam,black2025rtc}. This behavior is closely tied to their non-streaming inference procedure: at each replan, existing WAMs regenerate a complete action chunk from scratch through multi-step denoising and discard the unexecuted suffix of the previous plan. This procedure creates two coupled limitations. First, the combination of a large video-generation backbone, joint video–action processing, and iterative denoising makes WAM inference computationally expensive. Regenerating a complete action chunk at every replan further compounds this cost, resulting in substantial response latency and slower adaptation to new observations~\citep{lu2026faster,ma2026fasterwam,cai2026ahawam,li2026efficientwam}. Second, independently generated chunks may be inconsistent with previously executed actions, causing abrupt transitions at replanning boundaries~\citep{lu2026faster,hoeg2024streaming}. Existing methods typically improve either inference efficiency or cross-chunk action continuity, but rarely both within a unified framework.

\begin{figure}[t]
  \centering
  \includegraphics[width=\textwidth]{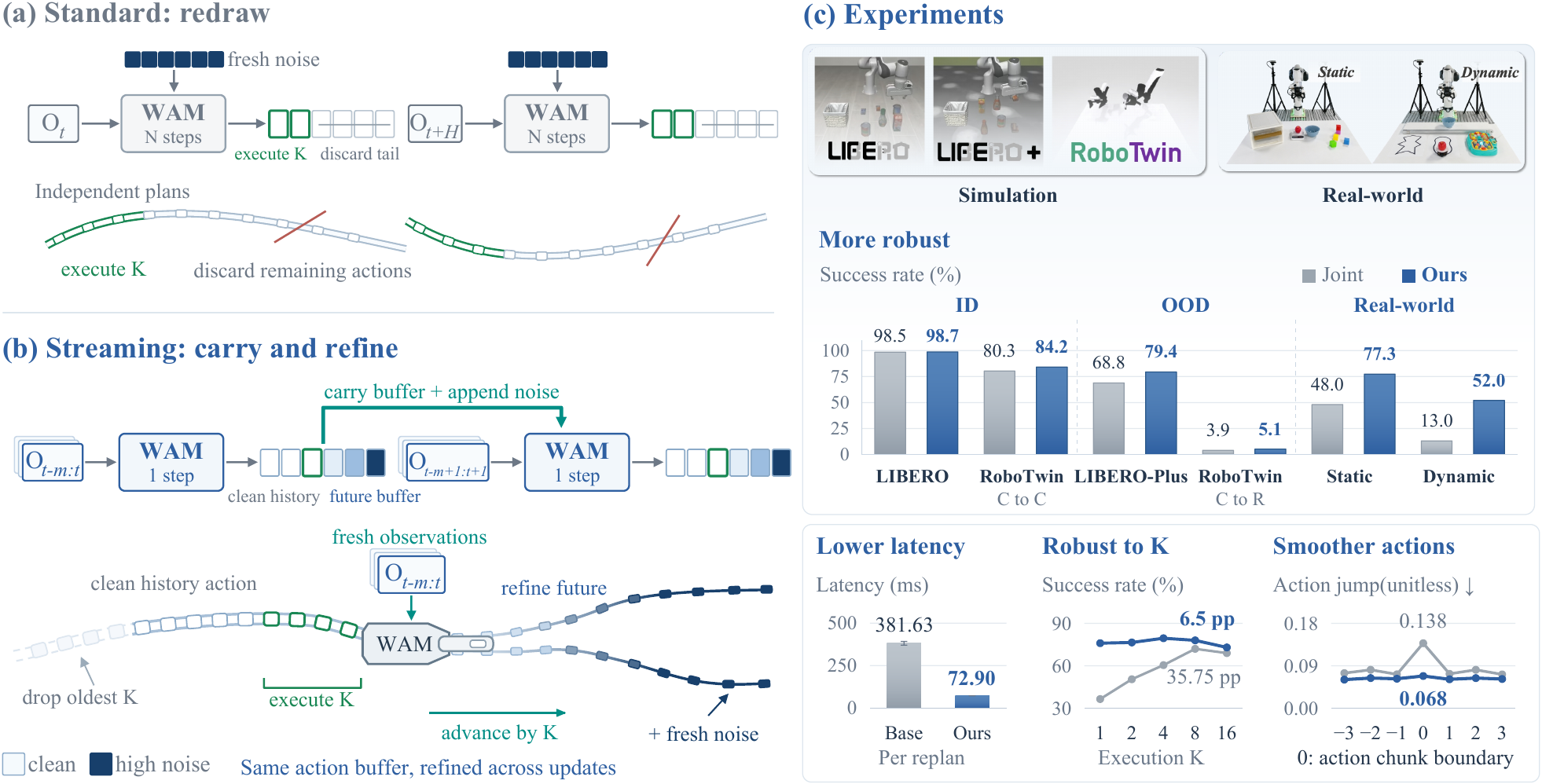}
  \caption{Overview of our streaming world-action model.
    (a)~Standard policies regenerate a complete action chunk at every replan.
    (b)~\method{} rolls the action buffer forward with one denoising step per replan.
    (c)~\method{} performs strongly across simulation and real-world benchmarks.}
  \label{fig:teaser}
\end{figure}

To bridge this gap, we propose \method{} (\cref{fig:teaser}), which introduces a streaming diffusion paradigm into WAMs. Specifically, the \textbf{noise ladder} staggers future action blocks across noise levels in a persistent buffer, enabling low-latency, single-step steady-state replanning. The \textbf{prefix rollout module} conditions future-action denoising on a clean prefix of recently executed actions, improving continuity across action chunks. Finally, \textbf{dense video-history memory} provides recent visual context for action prediction, supporting temporally consistent responses to state changes.

We evaluate \method{} on LIBERO, \liberoplus, RoboTwin~2.0, and real-world manipulation tasks spanning static and dynamic settings, where it achieves competitive success while reducing steady-state replanning latency to 72.9\,ms in simulation and 57.3\,ms on the real robot. Further analysis shows that the proposed components jointly improve cross-chunk action continuity, reduce sensitivity to the execution horizon~$\Kexec$, and support adaptation to changing observations.

Our contributions are summarized as follows:

\begin{itemize}\setlength{\itemsep}{2pt}\setlength{\parskip}{0pt}\setlength{\parsep}{0pt}
    \item We introduce \method{}, which integrates a noise ladder for chunk-wise streaming action prediction, a prefix rollout module for maintaining continuity with previously executed actions, and dense video-history memory for incorporating historical visual context into action prediction. Together, these components improve inference efficiency and execution continuity.
    \item We evaluate \method{} on LIBERO, \liberoplus, RoboTwin~2.0, and real-world manipulation tasks spanning both static and dynamic settings. \method{} achieves competitive task success while reducing steady-state replanning latency to 72.9\,ms in simulation and 57.3\,ms on the real robot. Extensive ablations further validate the contribution of each proposed component.
    \item Through controlled ablations and static and dynamic evaluations, we analyze
    the complementary effects of the proposed components. The noise ladder and
    prefix rollout improve cross-call action continuity, reduce sensitivity to the
    execution horizon~$\Kexec$, and enable low-latency replanning. The clean action
    prefix and dense video history provide recent action--observation context,
    supporting temporally consistent execution and adaptation to object motion.
\end{itemize}

\section{Related Work}
\label{sec:related}

\parhead{World action models and vision-language-action models}
Two dominant paradigms have emerged for generalist robot control.
Vision-language-action models (VLAs) such as
$\pi_0$~\citep{black2024pi0} and
$\pi_{0.5}$~\citep{black2025pi05} build on pretrained vision-language
models, combining Internet-scale semantic knowledge with large-scale
cross-embodiment robot training.
WAMs instead build on pretrained video-generation models and jointly
predict future visual states and actions, transferring video-derived
dynamics priors to robot control.
DreamZero~\citep{ye2026dreamzero} and
LingBot-VA~\citep{li2026lingbotva} demonstrated the viability of this
paradigm, achieving strong generalization across environments and
embodiments.
However, large video backbones and iterative joint video--action
denoising introduce substantial inference latency, limiting
high-frequency closed-loop control.

\parhead{Efficient world action models}
Recent work accelerates WAM inference through lightweight action modules,
reduced video-side computation, asynchronous video--action denoising, and
action-centered architectures~\citep{yuan2026fastwam,ma2026fasterwam,li2026efficientwam,guo2026xwam,ye2026gigaworldpolicy}.
These methods mainly reduce per-step computation while retaining the standard
full-chunk, multi-step replanning loop; in contrast, \method{} carries and
refines a persistent action buffer with a single forward pass per replan.

\parhead{Diffusion policies and streaming diffusion}
Diffusion Policy generates action chunks through iterative denoising~\citep{chi2023diffusionpolicy},
whereas Streaming Diffusion Policy (SDP) and Diffusion Forcing use position-
or token-dependent noise levels to support rolling and flexible-horizon
generation~\citep{hoeg2024streaming,chen2024diffusionforcing}.
\method{} adapts this principle to WAMs, preserving the video-conditioned
dynamics prior while enabling persistent, single-pass replanning.

\section{Method}
\label{sec:method}

\begin{figure*}[t]
\centering
\includegraphics[width=\textwidth]{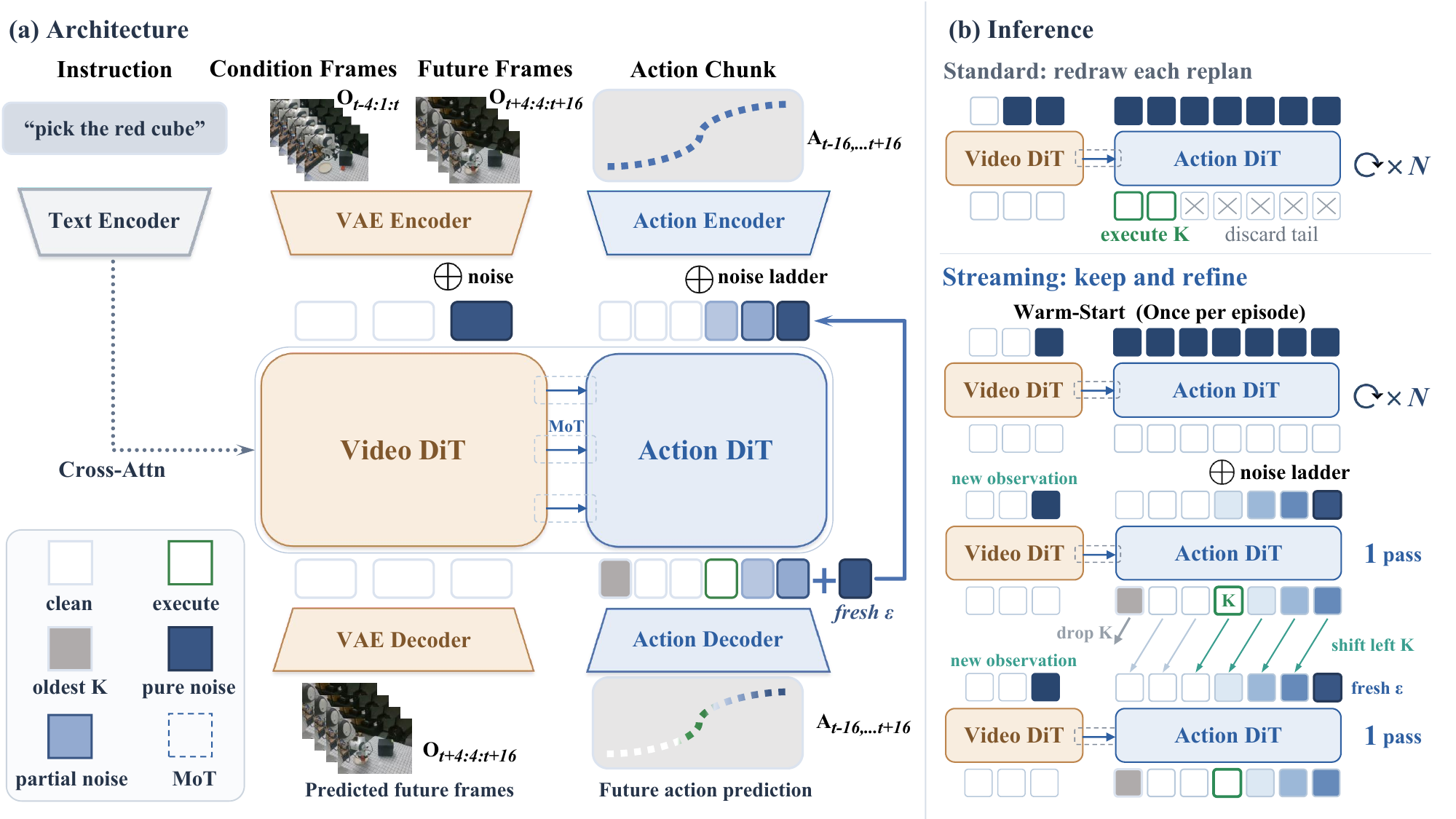}
\caption{Overview of \method{}. (a)~Joint video--action architecture with a clean action prefix and staircase denoising. (b)~Standard multi-step inference versus single-step streaming inference.}
\label{fig:method}
\end{figure*}

\subsection{Problem formulation}
\label{sec:method:formulation}

At each planning step \(t\), the robot queries the policy with the current
visual observation \(o_t\) and task instruction \(l\) to obtain an \(H\)-step
chunk of future actions:
\begin{equation}
\label{eq:policy}
    \act_{t:t+H-1}=\pi(\obs_t,l).
\end{equation}
Here, \(H\) denotes the future-action prediction horizon. A WAM realizes this
policy by jointly denoising the action chunk and a
short future video sequence \(v_{t:t+T}\), where \(T\) is the video prediction
horizon. At deployment, an execution horizon \(K \leq H\) determines how many
predicted actions are applied before replanning: the robot executes
\(a_{t:t+K-1}\), discards the unused suffix \(a_{t+K:t+H-1}\), and replans when
the next observation becomes available. For block-aligned streaming deployment,
we use execution widths that divide \(H\), ensuring that each roll advances the
buffer by an integer number of complete noise-ladder blocks.

\parhead{Streaming policy formulation.}
To enable streaming action control, we augment this standard WAM interface with
temporal context and a persistent plan buffer. At each
replanning time \(t\), the model consumes the \(m\) most recent visual
observations, the last \(\nprefix\) actions executed by the robot, and the
unfinished plan carried over from the previous call. It returns the next
\(\Kexec\) executable actions together with an updated plan buffer for the
subsequent call:
\begin{equation}
\label{eq:ours}
\left(
    \act_{t:t+\Kexec-1},
    \mathcal{B}_{t+\Kexec}
\right)
=
\pi_\theta\left(
    \underbrace{o_{t-m+1:t}}_{\text{recent visual observations}},
    \, l,\,
    \underbrace{\act_{t-\nprefix:t-1}}_{\text{executed-action history}},
    \underbrace{\mathcal{B}_t}_{\text{unfinished plan}}
\right).
\end{equation}
Here, \(t\) indexes environment steps, \(\nprefix\) is the executed-action
history length, and \(\mathcal{B}_t\) denotes the persistent plan buffer aligned
with the current control step. The robot executes only
\(\act_{t:t+\Kexec-1}\). Although the model maintains an \(H\)-step action plan
internally, action slots beyond the executable prefix remain in
\(\mathcal{B}_{t+\Kexec}\) and are carried forward to the next planning call.
These future actions may still be partially denoised and are never issued to the robot until they enter the executable prefix.

\parhead{Standard WAM inference lacks cross-call plan persistence.}
At the \(n\)-th planning call, a standard WAM initializes a new \(H\)-step
action chunk from independent noise and runs the complete denoising procedure:
\begin{equation}
\label{eq:redraw}
    \act^{(n)}_{1:H}
    =
    \mathcal{D}_\theta\!\left(
        \epsilon^{(n)}_{1:H};z_n
    \right),
    \qquad
    \epsilon^{(n)}_{1:H}\sim\mathcal{N}(0,I),
\end{equation}
where \(z_n=z(o_{t_n},l)\), \(t_n\) is the environment time of planning call
\(n\), and \(\mathcal{D}_\theta\) denotes the complete denoising procedure.
Neither the noise initialization nor the intermediate denoising state is passed
to call \(n+1\). After executing the first \(\Kexec\) actions, the unexecuted
suffix is discarded. Because all action slots follow the same denoising
schedule, the model neither reuses the previous plan nor maintains partially
refined future actions across calls. The streaming interface in
\Cref{eq:ours} therefore requires a persistent plan buffer and a
position-dependent denoising schedule.

\subsection{\texorpdfstring{\method}{Our method}}
\label{sec:method:model}

\parhead{Overview}
We propose \method{} (\cref{fig:method}), a streaming diffusion formulation built on a Mixture-of-Transformers (MoT) WAM with dedicated video and action experts. It combines a \textbf{prefix rollout module} that conditions prediction on clean executed actions, a \textbf{noise ladder} that progressively refines a persistent future-action buffer, and \textbf{dense video-history memory} that supplies recent visual context. Together, these components amortize denoising across planning calls, reducing steady-state latency while improving cross-chunk continuity. \method{} requires neither additional trainable parameters nor additional training data.

\parhead{Prefix rollout module and dense video-history memory}
At each planning step \(t\), we form an action sequence consisting of a clean prefix of \(\nprefix\) previously executed actions, followed by \(H\) future action slots. The prefix is not a prediction target: during training, it is populated with ground-truth actions preceding the target chunk, whereas at deployment, it contains the actions actually executed
by the robot. These prefix actions are provided without noise, allowing the
action branch to model the future action sequence as
\begin{equation}
\label{eq:prefix}
p_\theta\!\left(
    \act_{t:t+H-1}
    \mid
    z(o_{t-m+1:t},l),\,
    \act_{t-\nprefix:t-1}
\right).
\end{equation}
Here, \(o_{t-m+1:t}\) forms the dense video-history memory, capturing short-term visual dynamics that complement the executed-action prefix to improve the temporal consistency of action generation. As planning advances, newly executed actions are rolled into the conditioning
prefix, while the unexecuted portion of \(\mathcal{B}_t\) is shifted forward and reused. Thus, overlapping future actions are not regenerated from scratch, breaking the cross-call independence of standard WAM inference described in
\cref{sec:method:formulation}. Following \citet{black2025rtc}, the action prefix also provides longer-term context than the short visual-history window.

\parhead{Streaming denoising over the future segment}
The clean prefix alone does not carry the future plan across calls, since the \(H\) future action slots would still be reinitialized from pure noise. We therefore assign them a staircase of noise levels. The future slots are grouped into consecutive blocks of width \(\Kexec\), corresponding to the execution horizon, with all positions in each block sharing one noise level and later blocks assigned higher levels. Accordingly, after every \(\Kexec\) actions are executed, the prefix and the carried plan are rolled forward by exactly one block. Writing $c(j)=\lfloor (j-1)/\Kexec\rfloor$ for the block index of the $j$-th future position and $C=\lceil H/\Kexec\rceil$ for the number of blocks, the noise level at that position is
\begin{equation}
\label{eq:ladder}
\tau_j=\frac{1}{N}\left(
\Big\lfloor \frac{N\,(c(j)+1)}{C} \Big\rfloor-\delta \right),
\qquad
\delta\sim\mathcal{U}\{0,\dots,\lfloor N/C\rfloor-1\},
\end{equation}
where $N$ is the number of noise levels. We use $N\ge C$, which makes the
offset set nonempty; the implementation additionally clamps discrete noise
indices to the valid range for robustness. The random offset $\delta$ shifts
the whole staircase down, exposing the model to the partially denoised states
that occur midway through rolling execution.

\parhead{Multi-width staircase training.}
The staircase is entirely determined by the block width $\Kexec$, and we do not hold it fixed. For each training example, we independently sample the execution horizon \(\Kexec\) uniformly from a finite candidate set \(\mathcal{K}\) and construct the corresponding noise profile. We retain the standard shared-timestep objective for a fraction of training steps to preserve the model's multi-step denoising capability, which is required during the warm-up phase at the beginning of each episode, when the initial plan is generated from pure noise.

\parhead{Training objective}
Both the video and action branches are trained with weighted flow matching.
Given a clean target \(y\), Gaussian noise
\(\epsilon\sim\mathcal{N}(0,I)\), and a noise level \(\tau\in[0,1]\), we construct
the interpolant \(y_\tau=(1-\tau)y+\tau\epsilon\) and regress its velocity
\(\epsilon-y\):
\begin{equation}
\label{eq:fm}
\mathcal{L}_{vid}
=
\mathbb{E}_{y,\epsilon,\tau}
\left[
w(\tau)
\left\|
f_\theta(y_\tau,\tau,o,l)-(\epsilon-y)
\right\|_2^2
\right],
\end{equation}
where \(w(\tau)\) is inherited from the base model and emphasizes intermediate
noise levels. Under ordinary shared-timestep action training, all \(H\) future
slots use the same scalar \(\tau\), so the weight is shared across the suffix:
\begin{equation}
\label{eq:actshared}
\mathcal{L}^{\mathrm{shared}}_{\mathrm{act}}
=
\mathbb{E}
\left[
w(\tau)\frac{1}{H}
\sum_{j=1}^{H}
\left\|
u^{\mathrm{shared}}_j
-
\left(\epsilon_j-\act^{\mathrm{fut}}_j\right)
\right\|_2^2
\right].
\end{equation}
Here, \(u^{\mathrm{shared}}_j\) is the predicted velocity for future slot
\(j\) under the shared suffix timestep.
Our noise ladder replaces this scalar with a future-slot profile
\(\boldsymbol{\tau}=(\tau_1,\ldots,\tau_H)\) defined by
\cref{eq:ladder}. The action prefix \(\act^{\mathrm{pre}}\) is provided as clean
conditioning and excluded from the loss. The full token-wise timestep input is
\(\bar{\boldsymbol{\tau}}=(\mathbf{0}_{\nprefix},\boldsymbol{\tau})\). The
staircase action objective is
\begin{equation}
\label{eq:lact}
\mathcal{L}_{\mathrm{act}}
=
\mathbb{E}
\left[
\frac{1}{H}
\sum_{j=1}^{H}
w(\tau_j)
\left\|
u^{\mathrm{fut}}_j
-
\left(\epsilon_j-\act^{\mathrm{fut}}_j\right)
\right\|_2^2
\right],
\qquad
u^{\mathrm{fut}}
=
f_\theta\!\left(
\act^{\mathrm{pre}},
\act^{\mathrm{fut}}_{\boldsymbol{\tau}},
\bar{\boldsymbol{\tau}},
o,l
\right).
\end{equation}
Here, \(\act^{\mathrm{fut}}_j\) and \(u^{\mathrm{fut}}_j\) denote the \(j\)-th
future action slot and its predicted velocity, respectively. Since \(\tau_j\) varies
across noise-ladder blocks, \(w(\tau_j)\) produces a corresponding block-wise
loss profile. Compared with \cref{eq:actshared}, the position-dependent weight
therefore moves inside the sum over future slots. The video objective remains
unchanged, and the total loss is
\(\mathcal{L}=\mathcal{L}_{\mathrm{act}}
+\lambda\mathcal{L}_{\mathrm{vid}}\), where $\lambda$ is the relative video-loss
weight. We use equal video and action weights, i.e., $\lambda=1$, in all
experiments.

\subsection{Inference Process}
At deployment, the future plan is maintained in a persistent buffer \(\mathcal{B}_t\) of
\(H\) action slots. At each planning call, a single joint forward pass performs
one Euler update toward zero noise. Inference uses the deterministic
zero-offset ladder (i.e., \(\delta=0\) in \cref{eq:ladder}). For future slot
\(j\), we define the pre- and post-update noise levels and their difference as
\begin{equation}
\tau^{\mathrm{pre}}_j=
\frac{1}{N}\Big\lfloor\frac{N(c(j)+1)}{C}\Big\rfloor,
\qquad
\tau^{\mathrm{post}}_j=
\frac{1}{N}\Big\lfloor\frac{Nc(j)}{C}\Big\rfloor,
\qquad
\Delta_j=\tau^{\mathrm{pre}}_j-\tau^{\mathrm{post}}_j.
\end{equation}
For our evaluated settings, \(C\) divides \(N\), so
\(\Delta_j=1/C\) for every future slot. Writing
\(\boldsymbol{\tau}^{\mathrm{pre}}=(\tau^{\mathrm{pre}}_1,\ldots,
\tau^{\mathrm{pre}}_H)\) and
\(\boldsymbol{\Delta}=(\Delta_1,\ldots,\Delta_H)\), the buffer is updated and
rolled as
\begin{equation}
\label{eq:roll}
\begin{gathered}
\widetilde{\mathcal{B}}_t
=
\mathcal{B}_t
-
\boldsymbol{\Delta}\odot
f_\theta^{\mathrm{act}}\!\left(
\mathcal{B}_t,\boldsymbol{\tau}^{\mathrm{pre}},
o_{t-m+1:t},l,\act_{t-\nprefix:t-1}
\right),
\qquad
\act_{t:t+\Kexec-1}
=
\widetilde{\mathcal{B}}_{t;1:\Kexec},
\\[2pt]
\mathcal{B}_{t+\Kexec}
=
\operatorname{Concat}\!\left(
\widetilde{\mathcal{B}}_{t;\Kexec+1:H},
\epsilon^{\mathrm{new}}
\right).
\end{gathered}
\end{equation}
Here, \(\widetilde{\mathcal{B}}_t\) is the buffer after the Euler update, and
\(\epsilon^{\mathrm{new}}\sim\mathcal{N}(0,I)\) is a fresh block of
\(\Kexec\) noisy actions. The leading block reaches zero noise and is executed,
while the remaining \(H-\Kexec\) actions are retained and shifted forward. The
executed block is added to the clean action prefix for the next planning call
but is not stored in \(\mathcal{B}_{t+\Kexec}\).

\parhead{Warm-up initialization.}
At the beginning of each episode, we perform a one-time warm-up to construct
the initial persistent buffer \(\mathcal{B}_0\). Specifically, we first generate
a clean action chunk using the standard multi-step flow-matching sampler and
then re-noise its blocks according to the staircase, following the denoise
initialization strategy of \citet{hoeg2024streaming}. Since no executed-action
history is available at this point, the action prefix
\(\act^{\mathrm{pre}}\) is initialized with zeros. After this warm-up, each
planning call follows the prefix-rollout procedure and requires only a single
joint forward pass.

\section{Experiments}
\label{sec:exp}

\subsection{Setup}
\label{sec:exp:setup}

\noindent
\begin{minipage}[c]{0.56\textwidth}
\parhead{Benchmarks and baselines.}
We conduct experiments on LIBERO~\citep{liu2023libero} and
\liberoplus~\citep{fei2025liberoplus}. LIBERO comprises four 10-task suites
(\textsc{object}, \textsc{spatial}, \textsc{libero-10}, and \textsc{goal}),
while \liberoplus extends these tasks to 10{,}030 perturbed evaluation
instances, enabling systematic evaluation under distribution shifts. We further
evaluate on \robotwin~\citep{chen2025robotwin2}, a dual-arm manipulation
benchmark with clean and domain-randomized settings, to assess whether the
findings transfer across benchmarks and embodiments.
\end{minipage}\hfill
\begin{minipage}[c]{0.40\textwidth}
\centering
\includegraphics[width=\linewidth]{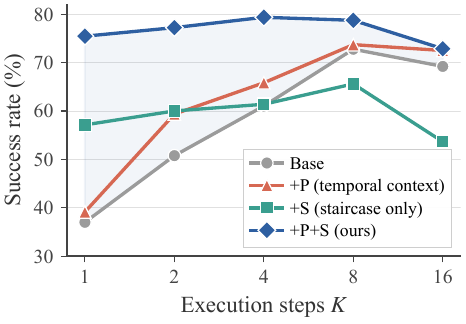}
\captionof{figure}{Execution-horizon sweep on the full \liberoplus evaluation.}
\label{fig:kcurve}
\end{minipage}
\par\vspace{0.25em}
\begin{minipage}{\textwidth}
\noindent Finally, we conduct experiments on both static and dynamic real-world manipulation tasks using a Franka Research~3 arm equipped with a Robotiq 2F-85 gripper, two external Intel RealSense D435i cameras, and a wrist-mounted ZED Mini camera (\cref{sec:exp:real}).
We compare \method{} against 13 competitive baselines spanning both VLA policies and WAMs, grouped according to whether they use embodied data during pretraining. Detailed baseline descriptions are provided in \cref{app:baselines}.
\end{minipage}
\parhead{Metrics.}
For most tasks, we report success rate~(\%); for Whack-a-Mole, we report valid
activations of the currently lit button within a fixed time budget. We
distinguish the once-per-episode time to first action~(TTFA), which includes
buffer initialization, from the steady-state replanning latency used during
normal closed-loop control~\citep{lu2026faster}.
Continuity metrics include per-step position change, joint displacement from
the initial posture, and a discrete acceleration proxy.

\parhead{Implementation details}
Following \fastwam~\citep{yuan2026fastwam}, we build our WAM on
Wan2.2-TI2V-5B. The action-expert backbone is initialized by linearly
interpolating the pretrained video-DiT weights to its dimensions and applying
\(\alpha\)-scaling to preserve their magnitude, while the action-specific
encoder and output head are randomly initialized. At inference, \(\mathcal{B}_0\) is initialized using a 10-step flow-matching sampler and re-noised according to the staircase. Each subsequent planning call uses a single Euler update. Detailed training and inference configurations, including the number of optimization and denoising steps, action and execution
horizons, are provided in \cref{app:repro}.
For controlled training-budget comparisons, \fwjoint and \method{} start from
the same benchmark-specific Joint checkpoint and receive the same number of
additional optimisation steps under the Joint and streaming objectives,
respectively. The main-table \liberoplus baseline instead follows the official
FastWAM-Joint evaluation configuration, as stated in the table caption.

\begin{table}[h!]
\centering
\caption{Success rates (\%) on LIBERO and \liberoplus. The final column reports
success pooled over all \liberoplus evaluation instances. FastWAM-Joint follows
its official evaluation configuration; detailed inference settings are provided
in \cref{tab:inference_config}.
  EP: embodied pre-training.
  \textbf{Bold}: best per column; \underline{underline}: second best.}
\label{tab:main}
\setlength{\tabcolsep}{3.0pt}
\footnotesize
\begin{tabular*}{\linewidth}{@{\extracolsep{\fill}}lcccccccccc@{}}
\toprule
  & & & \multicolumn{7}{c}{Perturbation type} & \\
  \cmidrule(lr){4-10}
Method & EP & LIBERO & Cam. & Robot & Lang. & Light & Bkgnd. & Noise & Layout & Overall \\
\midrule
\multicolumn{11}{@{}l}{\textit{With embodied pre-training}} \\
$\pi$0.5      & \checkmark & 96.9 & \textbf{75.4} & 77.5 & 85.6 & \underline{96.9} & \underline{94.6} & \underline{89.7} & \textbf{85.7} & \textbf{85.7} \\
GE-ACT        & \checkmark & 96.5 & 60.7 & 77.0 & 77.4 & 95.8 & 86.0 & \textbf{90.9} & 80.2 & \underline{80.3} \\
VLA-JEPA      & \checkmark & 97.2 & 63.3 & 67.1 & 85.4 & 95.6 & 93.6 & 66.3 & \underline{85.1} & 79.5 \\
X-VLA         & \checkmark & 98.1 & 23.4 & \textbf{89.7} & 75.7 & 88.2 & \textbf{96.0} & 62.7 & 71.8 & 71.4 \\
OpenVLA-OFT   & \checkmark & 97.1 & 56.4 & 31.9 & 79.5 & 88.7 & 93.3 & 75.8 & 74.2 & 69.6 \\
$\pi$0-Fast   & \checkmark & 85.5 & 65.1 & 21.6 & 61.0 & 73.2 & 73.2 & 74.4 & 68.8 & 61.6 \\
$\pi$0        & \checkmark & 94.2 & 13.8 & 6.0 & 58.8 & 85.0 & 81.4 & 79.0 & 68.9 & 53.6 \\
UniVLA        & \checkmark & 95.2 & 1.8 & 46.2 & 69.6 & 69.0 & 81.0 & 21.2 & 31.9 & 42.9 \\
\midrule
\multicolumn{11}{@{}l}{\textit{Without embodied pre-training}} \\
\fasterwam    & -- & 98.0 & \underline{67.9} & 49.0 & \underline{92.1} & 94.3 & 57.0 & 82.3 & 82.7 & 75.0 \\
4D-WAM        & -- & 98.4 & 45.2 & 64.3 & 90.6 & 94.3 & 57.7 & 69.1 & 79.2 & 71.0 \\
\fwjoint & -- & \underline{98.5} & 42.5 & 60.8 & \textbf{92.9} & 91.8 & 55.4 & 63.3 & 78.4 & 68.8 \\
WorldVLA      & -- & 79.1 & 0.1 & 27.9 & 41.6 & 43.7 & 17.1 & 10.9 & 38.0 & 25.0 \\
\fastwam      & -- & 97.6 & 16.4 & 44.5 & 68.9 & 78.2 & 53.7 & 37.7 & 60.7 & 51.5 \\
\midrule
\textbf{Ours}
  & -- & \textbf{98.7} & 61.0 & \underline{79.0} & 92.0 & \textbf{97.8} & 64.7 & 78.6 & 83.6 & 79.4 \\
\bottomrule
\end{tabular*}
\end{table}

\subsection{Main results}
\label{sec:exp:main}
\label{sec:exp:robotwin}

\Cref{tab:main} shows that \method{} achieves the strongest overall performance among WAMs without embodied-data pretraining, reaching 98.7\% on LIBERO and 79.4\% on \liberoplus. The gain is most pronounced under distribution shift, where it outperforms the closest architectural baseline, \fwjoint, by 10.6 percentage points on \liberoplus, particularly under camera-viewpoint and robot-initial-state perturbations. This comparison uses each method's reported deployment configuration (\cref{tab:inference_config}) and is a system-level comparison rather than a matched-$\Kexec$ ablation.

\begin{wraptable}[16]{r}{0.54\textwidth}
\vspace{-0.8em}
\centering
\captionsetup{hypcap=false,width=\linewidth}
\caption{RoboTwin~2.0 success rate (\%). EP: embodied pre-training;
bold/underline: best/second best.}
\label{tab:robotwin}
\begingroup\footnotesize
\renewcommand{\arraystretch}{0.94}
\begin{tabular*}{\linewidth}{@{\extracolsep{\fill}}lcccc@{}}
\toprule
Method & EP & Clean & Random & Avg. \\
\midrule
\multicolumn{5}{@{}l}{\textit{With embodied pre-training}} \\
Spatial Forcing             & \checkmark & 77.2 & 9.5  & 43.4 \\
$\pi$0.5                    & \checkmark & 70.7 & \textbf{46.0} & \textbf{58.4} \\
X-WAM                       & \checkmark & 70.0 & \underline{25.8} & \underline{47.9} \\
X-VLA                       & \checkmark & 68.0 & 20.9 & 44.5 \\
EventVLA                    & \checkmark & 65.6 & 15.7 & 40.7 \\
Xiaomi-Robotics-0           & \checkmark & 62.9 & 18.2 & 40.6 \\
G0                          & \checkmark & 62.7 & 9.1  & 35.9 \\
Abot-M0                     & \checkmark & 57.4 & 22.9 & 40.2 \\
\midrule
\multicolumn{5}{@{}l}{\textit{Without embodied pre-training}} \\
\fwjoint                    & --         & \underline{80.3} & 3.9  & 42.1 \\
\fastwam                    & --         & 77.8 & 1.9  & 39.9 \\
AHA-WAM                     & --         & 64.3 & 3.2  & 33.8 \\
\midrule
\textbf{Ours}               & --         & \textbf{84.2} & 5.1  & 44.7 \\
\bottomrule
\end{tabular*}\endgroup
\vspace{-0.5em}
\end{wraptable}

For RoboTwin~2.0 (\cref{tab:robotwin}), task-specific training uses the same 2{,}500 clean demonstrations. \method{} reaches 84.2\% clean and 5.1\% randomized success, compared with 80.3\% and 3.9\% for \fwjoint. The gap to embodied-pretrained \(\pi_{0.5}\) highlights the value of cross-embodiment data for out-of-distribution generalization.

\par
\begin{figure*}[ht!]
\centering
\begin{subfigure}[t]{0.245\textwidth}
  \centering
  \includegraphics[width=\linewidth]{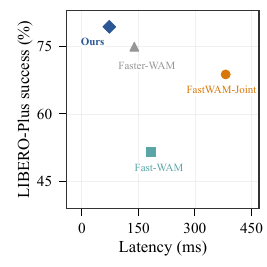}
  \caption{Simulation efficiency.}
  \label{fig:latency}
\end{subfigure}\hfill
\begin{subfigure}[t]{0.245\textwidth}
  \centering
  \includegraphics[width=\linewidth]{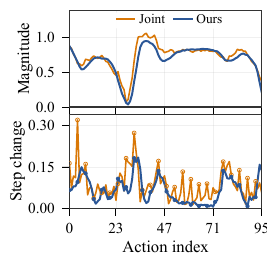}
  \caption{Simulation continuity.}
  \label{fig:smoothness}
\end{subfigure}\hfill
\begin{subfigure}[t]{0.245\textwidth}
  \centering
  \includegraphics[width=\linewidth]{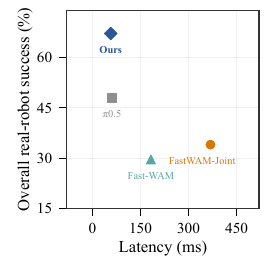}
  \caption{Real-robot efficiency.}
\end{subfigure}\hfill
\begin{subfigure}[t]{0.245\textwidth}
  \centering
  \includegraphics[width=\linewidth]{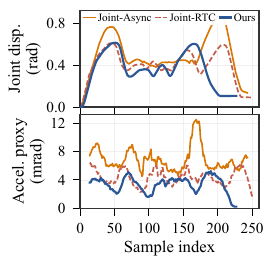}
  \caption{Real-robot continuity.}
\end{subfigure}
\caption{Inference efficiency and action continuity in simulation and on
  the real robot. Panel~(c) is a system-level summary: for Fast-WAM and
  FastWAM-Joint,
  the latency coordinate is the standard 10-step latency, while overall success
  uses the task-specific real-robot deployment in which dynamic tasks use one
  denoising step.}
\label{fig:latency_smoothness}
\end{figure*}

\medskip
\noindent\textbf{Inference cost}\hspace{0.4em}
We report episode-initial TTFA separately from steady-state replanning latency. \method{} takes 72.9\,ms in simulation and 57.3\,ms on the real robot during normal steady-state control (\cref{fig:latency_smoothness}), corresponding to $5.2\times$ and $6.4\times$ speedups over the standard 10-step \fwjoint deployment. These system-level comparisons include different horizons and evaluation counts. One real-robot update is 57.3\,ms versus 46.8\,ms for a timing-only one-step Joint call; the speedup comes from amortizing denoising so each steady replan uses one model evaluation instead of ten. The one-time episode-initial warm start is reported separately in \cref{tab:latency_detail}.

\par\medskip
\noindent\textbf{Action continuity}\hspace{0.4em}
\method{} produces smoother actions in both simulation and real-world execution. On \liberoplus, the boundary-to-interior jump ratio drops from 1.840 to 1.105 and global translation-command jerk decreases by 59.0\% relative to \fwjoint (\cref{fig:smoothness,tab:smoothness}).

\par\vspace{0.35em}
\noindent Real-robot trajectories show the same trend, with smoother joint displacement and a lower discrete-acceleration proxy (\cref{fig:latency_smoothness}(d)).
\par

\subsection{Real robot experiments}
\label{sec:exp:real}

\begin{figure*}[t]
\centering
\begin{minipage}{\textwidth}
\centering
\begin{subfigure}[t]{0.495\linewidth}
  \centering
  \includegraphics[width=\linewidth]{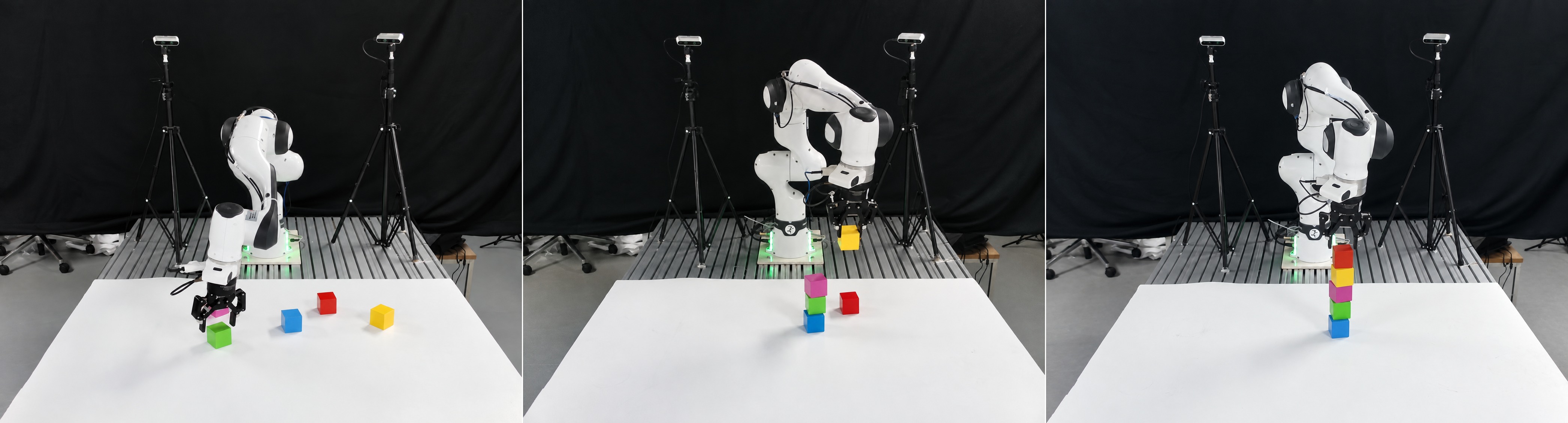}
  \caption{Block stacking}
\end{subfigure}\hfill
\begin{subfigure}[t]{0.495\linewidth}
  \centering
  \includegraphics[width=\linewidth]{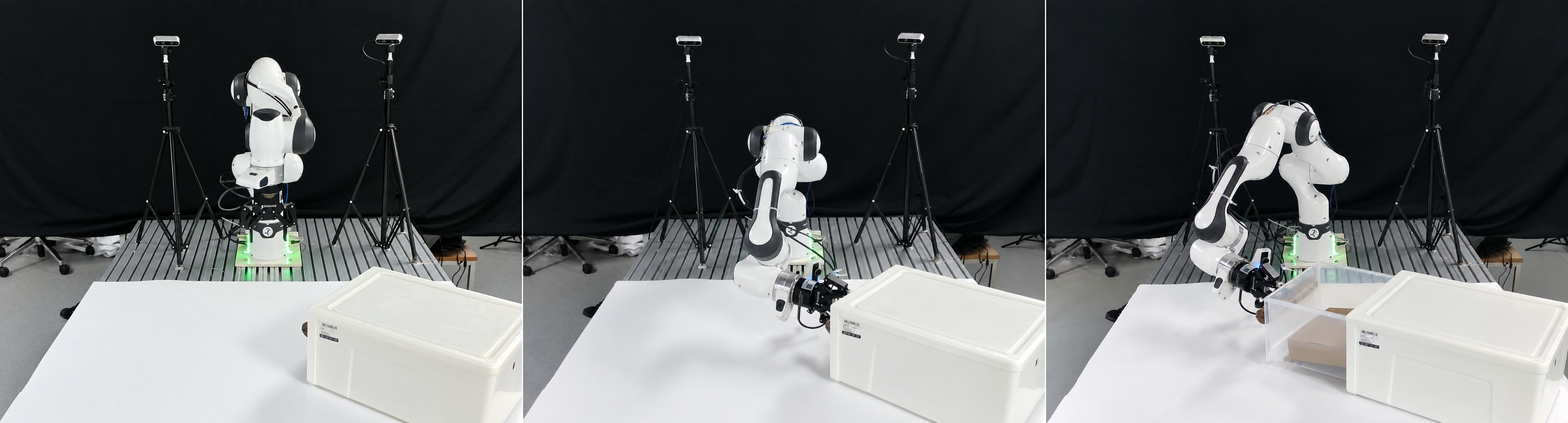}
  \caption{Open drawer}
\end{subfigure}

\par\vspace{2pt}

\begin{subfigure}[t]{0.495\linewidth}
  \centering
  \includegraphics[width=\linewidth]{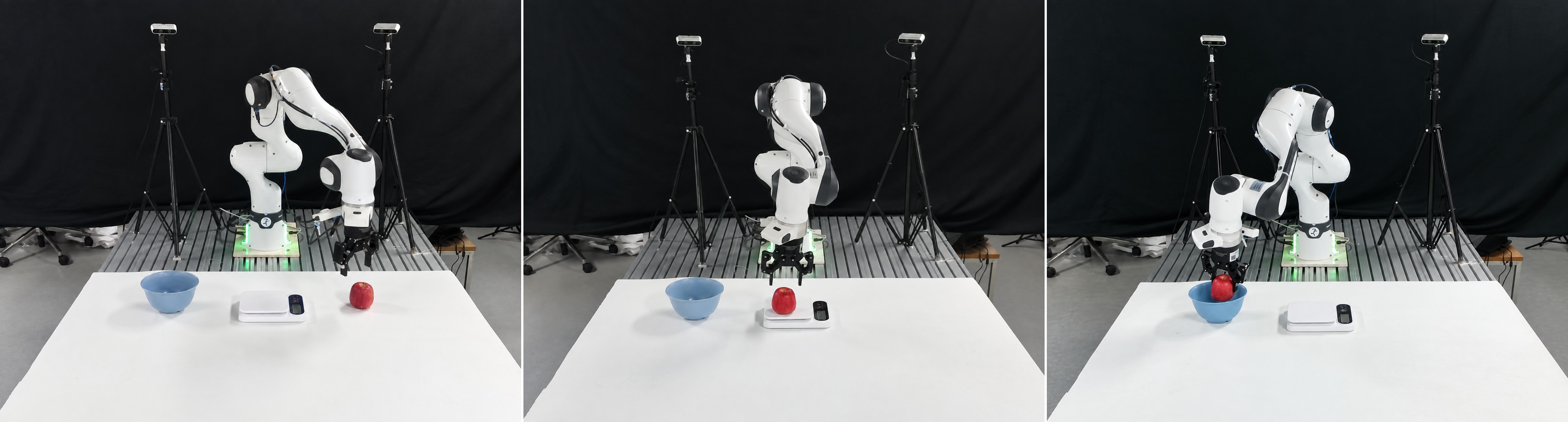}
  \caption{Weigh-and-place}
\end{subfigure}\hfill
\begin{subfigure}[t]{0.495\linewidth}
  \centering
  \includegraphics[width=\linewidth]{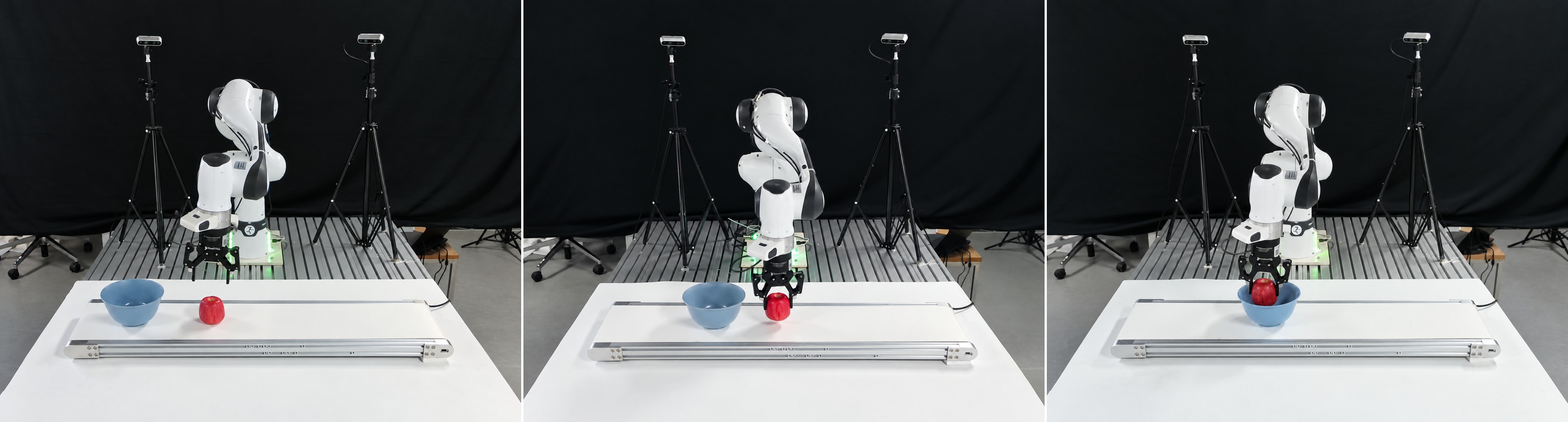}
  \caption{Conveyor handover}
\end{subfigure}

\par\vspace{2pt}

\begin{subfigure}[t]{0.495\linewidth}
  \centering
  \includegraphics[width=\linewidth]{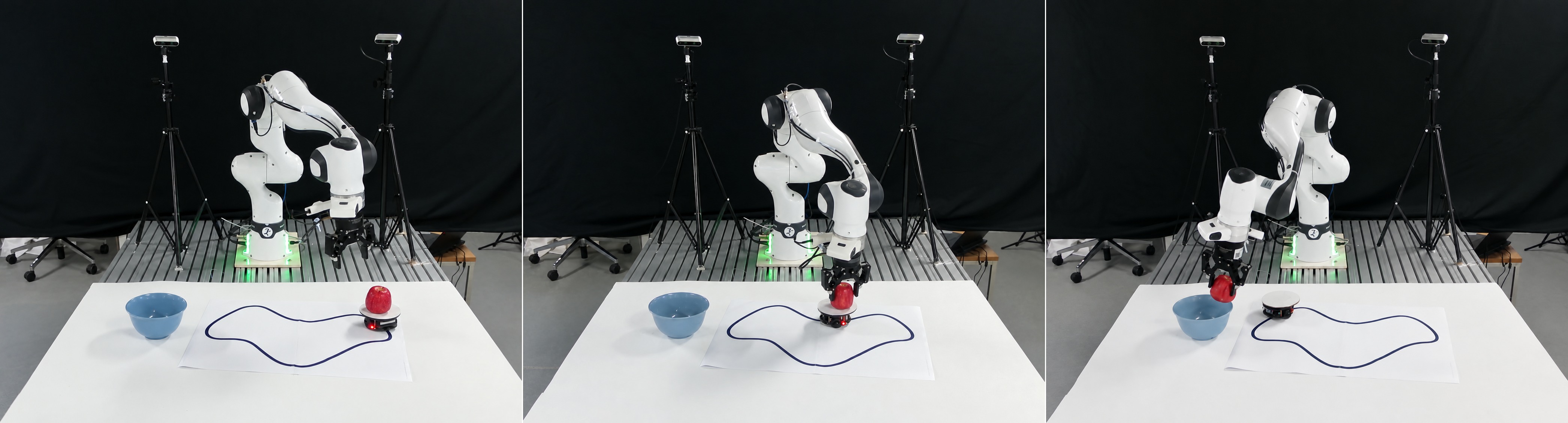}
  \caption{Car pursuit}
\end{subfigure}\hfill
\begin{subfigure}[t]{0.495\linewidth}
  \centering
  \includegraphics[width=\linewidth]{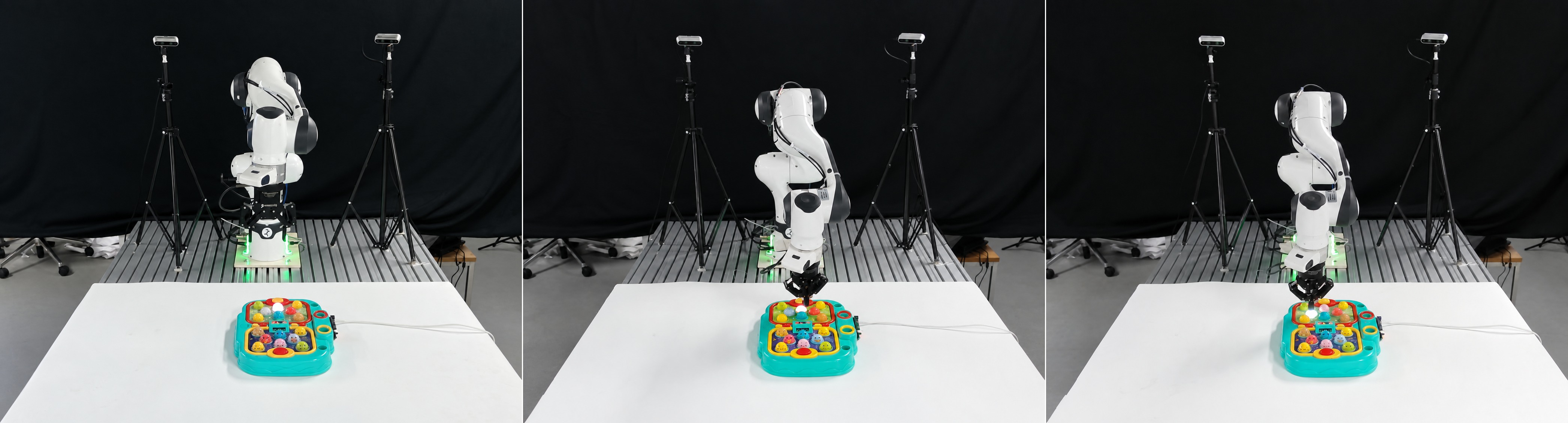}
  \caption{Whack-a-mole}
\end{subfigure}
\end{minipage}

\caption{Real-robot tasks. (a)--(c) Static; (d)--(f) dynamic. Each panel shows
one successful rollout.}
\label{fig:real_tasks}
\end{figure*}

\parhead{Setup}
We deploy on a Franka Research~3 arm with a Robotiq 2F-85 gripper, two external
Intel RealSense D435i cameras, and a wrist-mounted ZED Mini following the DROID~\citep{khazatsky2024droid}
configuration. The control loop runs at 15\,Hz on an RTX~5090 workstation.
Demonstrations are collected via GELLO joint-space teleoperation; details are
provided in \cref{app:real_tasks}.
On the deployment GPU, steady-state per-replan latency is
57.3\,ms for our model (1 denoise step) and 61.5\,ms for $\pi_{0.5}$
(10 steps), both within the 66.7\,ms control budget.
Fast-WAM and FastWAM-Joint require 182.6\,ms and 368.6\,ms, respectively, with
their standard 10-step inference, missing multiple deadlines per replan
(\cref{tab:latency_detail}). We therefore use 10 steps on static tasks and one
step on dynamic tasks for Fast-WAM and FastWAM-Joint; preliminary
10-step dynamic deployments yielded near-zero success because the policy could
not react within the task time scale.
The six evaluation tasks are illustrated in \cref{fig:real_tasks}, with detailed
descriptions provided in \cref{app:real_task_descriptions}.

\begin{figure}[H]
\centering
\includegraphics[width=\linewidth]{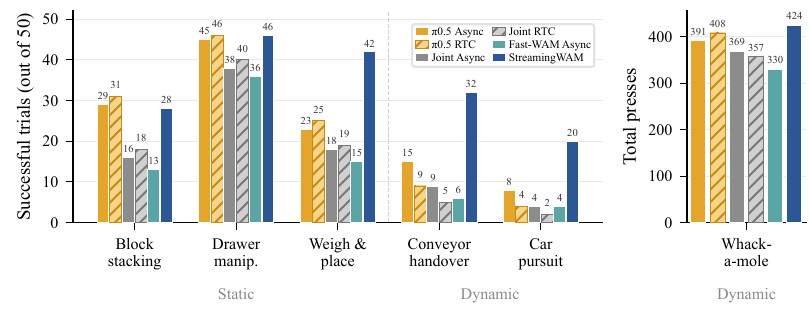}
\caption{Real-robot results.  Each of the five manipulation tasks
  (left) reports the number of successful trials out of 50; whack-a-mole (right) reports
  total valid lit-button activations across 50$\times$30\,s sessions.}
\label{fig:real_results}
\end{figure}

\parhead{Results}
\Cref{fig:real_results} shows that \method{} performs best overall.
It achieves the highest aggregate success across the five manipulation tasks
(168/250, 67.2\%, Wilson 95\% CI [61.2, 72.7], versus 120/250, 48.0\% [41.9, 54.2], for the strongest baseline), leads both dynamic tasks
(32/50 on conveyor handover and 20/50 on car pursuit, compared with baseline
maxima of 15/50 and 8/50), and records the most whack-a-mole presses
(424 vs. 408).

\parhead{Analysis}
Two complementary temporal effects may explain the task-level gains. First,
the largest margins occur on the dynamic tasks. A single frame locates an
object but generally leaves its velocity ambiguous, whereas consecutive real
observations reveal frame-to-frame motion. This temporal evidence may allow the
WAM's future-state predictor to anticipate object motion before action
selection. At the same time, single-step streaming replanning lowers the
steady-state latency, allowing those predictions to be refreshed at the task's
reaction time scale. The dynamic gains therefore plausibly reflect the joint
effect of motion evidence from real observation history, the WAM's predictive
prior, and low-latency replanning, rather than any one factor alone. In
stage-matched conveyor trials, FastWAM-Joint largely freezes the
apple, whereas \method{} predicts displacement along the conveyor direction.
The corresponding qualitative future-video prediction sequence and displacement
measurements are provided in the appendix (\cref{fig:real_motion_anticipation}). Because the
examples come from independent trials and the models differ in more than their
visual context, they provide qualitative mechanism evidence rather than a
controlled history-only ablation. Second, temporal context may also help track
progress in multi-stage tasks. On weigh-and-place, \method{} succeeds in 42/50
trials versus 25/50 for the strongest baseline. Recent observations together
with the executed-action history provide short-horizon context about whether
the weighing stage has been completed and whether the apple should next be
transferred to the bowl. This interpretation is consistent with the task-level
gain, although it remains to be isolated by a task-specific context ablation.

\subsection{Ablation studies and analysis}
In this section, we ablate the key components of \method{} and conduct further
analyses of its behavior, yielding the following findings:
\label{sec:exp:ablation}

\begin{table}[t]
\centering
\caption{Ablation studies on the full 10{,}030-episode \liberoplus
  evaluation (success rate, \%).
  The top block compares the temporal-context package~(P) and streaming
  staircase~(S); the bottom block varies one design choice from the full
  model. Spread is max$-$min across the reported $\Kexec$ values.}
\label{tab:ablation}
\setlength{\tabcolsep}{3.6pt}
\footnotesize
\begin{tabular*}{\linewidth}{@{\extracolsep{\fill}}lcccccr@{}}
\toprule
Variant & $K{=}1$ & $2$ & $4$ & $8$ & $16$ & Spread \\
\midrule
Base (\fwjoint) & 37.01 & 50.77 & 61.03 & 72.76 & 69.21 & 35.75 \\
+ Temporal context (P) & 39.14 & 59.36 & 65.83 & 73.71 & 72.49 & 34.57 \\
+ Staircase (S) & 57.13 & 60.01 & 61.39 & 65.60 & 53.65 & 11.95 \\
+ P + S (full model) & \textbf{75.45} & \textbf{77.22} & \textbf{79.36} & \textbf{78.73} & \textbf{72.86} & \textbf{6.50} \\
\midrule
\quad w/ single-width $K{=}4$ & 71.52 & 74.08 & 75.31 & 73.74 & 70.39 & 4.92 \\
\quad w/ sparse history & 56.35 & 67.80 & 69.86 & 68.45 & 60.43 & 13.51 \\
\quad w/ uniform loss weight & 75.52 & 78.37 & 78.11 & 77.50 & 71.27 & 7.10 \\
\quad w/o warm start & 70.30 & 72.86 & 74.21 & 75.23 & 72.56 & 4.93 \\
\bottomrule
\end{tabular*}
\end{table}

\parhead{(1) The staircase flattens the performance curve across execution
horizons, and the combined temporal-context package raises its overall level.}
\Cref{fig:kcurve,tab:ablation} compares the clean action prefix bundled with dense visual history (P) and the streaming
staircase (S) across execution horizons.  The staircase reduces Spread from
35.75\,pp for Base and 34.57\,pp for +P to 11.95\,pp for +S; combining P and S
further reduces it to 6.50\,pp while achieving the highest success at every
evaluated \(\Kexec\). Thus S provides \(\Kexec\)-robustness, while the bundled
temporal context raises performance; finding~(3) separately tests history density.

\parhead{(2) Multi-width staircase training improves performance across execution
horizons.}
As shown in \cref{tab:ablation}, replacing the full training set
\(\mathcal{K}\) with a fixed staircase width of \(\Kexec=4\) preserves low
variation across execution horizons but reduces success at every evaluated
value of \(\Kexec\), including its training width (75.31\% versus 79.36\% at
\(\Kexec=4\)). The fixed-width model peaks at \(\Kexec=4\), as expected, but
generalizes less effectively to other execution horizons. These results support the design motivation in
\cref{sec:method:model}: sampling multiple widths exposes the model to more
diverse staircase geometries and improves performance across the entire
\(\Kexec\)-sweep.

\parhead{(3) Dense visual history improves success rates and reduces variation across
execution horizons.}
With P and S held fixed, replacing dense visual history (stride~1) with sparse
history (stride~4) shifts the entire \(\Kexec\)-sweep downward and nearly
doubles the Spread (\cref{tab:ablation}).
The dense frames supply recent visual context that overlaps with the
clean-prefix window, giving the model a more accurate estimate of the
current scene state.
The benefit is most pronounced at small \(\Kexec\), where frequent replanning places greater
importance on recent observations.

\parhead{(4) Auxiliary training choices mainly affect accuracy rather than
horizon robustness.}
Uniform loss weighting stays within 1.59\,pp of the full model across all
\(\Kexec\) and changes Spread only from 6.50 to
7.10\,pp, indicating that position-dependent weighting is a modest refinement.
On the full evaluation, removing the warm start reduces the $K=4$ result by
5.15\,pp (74.21\% versus 79.36\%). This shows that initialization mainly raises
the initial buffer's accuracy; the staircase remains the mechanism tested for
robustness across execution horizons.

\parhead{(5) The streaming advantage vanishes when the execution width spans
the full prediction horizon.}
The training noise profiles sample
\(\mathcal{K}_{\mathrm{train}}=\{1,2,4,8\}\), all block-aligned divisors of
\(H=16\), while \(\Kexec=16\) is held out from training. At
\(\Kexec=H=16\), the horizon becomes one block and
loses the inter-block staggering required for streaming. The full model then
drops by 5.87\,pp from \(\Kexec=8\) to \(16\) (78.73\% to 72.86\%), while its
margin over the temporal-context-only (+P) variant shrinks from 13.53\,pp at \(\Kexec=4\) to
0.37\,pp at \(\Kexec=16\). Thus, this boundary stress test shows that the
streaming benefit vanishes when the execution width spans the prediction
horizon.

\section{Limitations}
\label{sec:limitations}

Our study has two main limitations. First, our experiments use a single
base-model family, \fastwam. Although the streaming formulation does not
require a specific WAM backbone, its effectiveness on other backbones remains
to be validated. Second, our model does not use large-scale embodied
pre-training. On RoboTwin~2.0's randomized-initialization split,
embodied-pretrained policies such as $\pi_{0.5}$~\citep{black2025pi05} retain
substantially higher success under scene variation. This gap may partly reflect
the broader data coverage and stronger generalization provided by
cross-embodiment pre-training. Combining such pre-training with streaming
denoising is an important direction for future work.

\section{Conclusion}
\label{sec:conclusion}

We presented \method{}, a streaming-diffusion framework that amortizes action
denoising across planning calls through a persistent rolling buffer. Its
streaming staircase progressively refines future actions, while the
executed-action prefix and dense visual history maintain temporal context
across replans. Across LIBERO-Plus, RoboTwin~2.0, and real-robot tasks,
\method{} achieves competitive task success, reduces steady-state replanning
latency to 72.9\,ms in simulation and 57.3\,ms on the real robot, improves
cross-chunk action continuity, and compresses execution-horizon sensitivity
from 35.75 to 6.50 percentage points. Its largest real-robot gains appear on
moving-target and multi-stage tasks, consistent with the value of temporal
context and responsive replanning. Together, these results support streaming
action generation as a practical approach to more responsive and continuous
WAM control.

\subsection*{Code and data availability}
Code, model checkpoints, the real-world dataset, and supplementary videos are
available from the \href{https://renshaojie233.github.io/StreamingWAM/}{project page}.

\bibliography{refs}
\bibliographystyle{plainnat}

\clearpage
\appendix
\section{Reproducibility details}
\label{app:repro}

\subsection{Model configuration}
\label{app:model_config}

We use the same dual-expert Mixture-of-Transformers backbone in all three
domains.  The Video and Action DiTs contain 30 aligned blocks.  In every
block, the two experts form their own queries, keys, and values, perform
joint mixed self-attention, and then return to expert-specific
cross-attention and feed-forward paths.  The Action DiT residual width is
1{,}024, whereas its 24 heads use an internal QKV width of
$24\!\times\!128=3{,}072$; these are different quantities.

\begin{table*}[h!]
\centering
\caption{Core model configuration by benchmark. Video/Action values are shown as V/A.}
\label{tab:model_config}
\setlength{\tabcolsep}{3.5pt}
\scriptsize
\begin{tabular*}{\textwidth}{@{\extracolsep{\fill}}lccccccc@{}}
\toprule
Benchmark & Layers & Hidden dim. & FFN dim. & Heads $\times$ dim. & Act./prop. & Cameras & Resolution ($H\!\times\!W$) \\
\midrule
LIBERO / \liberoplus & 30/30 & 3{,}072/1{,}024 & 14{,}336/4{,}096 & $24\!\times\!128$ & 7/8 & 2, horiz. & $224\!\times\!448$ \\
RoboTwin~2.0 & 30/30 & 3{,}072/1{,}024 & 14{,}336/4{,}096 & $24\!\times\!128$ & 14/14 & 3, top + 2 bottom & $384\!\times\!320$ \\
Real robot & 30/30 & 3{,}072/1{,}024 & 14{,}336/4{,}096 & $24\!\times\!128$ & 8/8 & 3, vertical & $384\!\times\!224$ \\
\bottomrule
\end{tabular*}
\end{table*}

\subsection{Training configuration}
\label{app:training_config}

For each direct comparison, the baseline and \method{} share a Joint
initialisation and receive the same number of additional optimisation steps.
The baseline retains ordinary shared-timestep Joint training, whereas
\method{} uses the streaming objective.  The resulting schedules are listed
in \cref{tab:training_config}.

For LIBERO and \liberoplus, the second-stage branches use AdamW with learning
rate $10^{-4}$, betas $(0.9,0.95)$, $\epsilon=10^{-8}$, weight decay
$10^{-2}$, gradient clipping at 1.0, bf16 precision, and a 60{,}000-step cosine
schedule without warm-up.  We report the checkpoint at step 40{,}000, selected
for evaluation.
For \method{}, 80\% of optimization minibatches use the staircase objective with
$\mathcal{K}_{\mathrm{train}}=\{1,2,4,8\}$, 32 noise levels, and a random
staircase offset; the remaining 20\% use ordinary shared-timestep flow
matching.  The video and action objectives have equal weight, the clean
16-step action prefix is conditioning only, and the future horizon is $H=16$.
We additionally evaluate the held-out, block-aligned width $K=16$.

\begin{table}[H]
\centering
\caption{Training schedule by benchmark.}
\label{tab:training_config}
\setlength{\tabcolsep}{10pt}
\small
\resizebox{\textwidth}{!}{%
\begin{tabular}{@{}llcccc@{}}
\toprule
Benchmark & Method & Shared Joint stage & Second stage & Total steps & Global batch \\
\midrule
LIBERO / \liberoplus
  & \fwjoint & 34{,}720 & $+$40{,}000 Joint & 74{,}720 & $80{\rightarrow}64$ \\
  & \method{} & 34{,}720 & $+$40{,}000 streaming & 74{,}720 & $80{\rightarrow}64$ \\
\midrule
RoboTwin~2.0
  & \fwjoint & 135{,}776 & $+$135{,}776 Joint & 271{,}552 & $64{\rightarrow}64$ \\
  & \method{} & 135{,}776 & $+$135{,}776 streaming & 271{,}552 & $64{\rightarrow}64$ \\
\midrule
Real robot
  & \fwjoint & 20{,}000 & $+$30{,}000 Joint & 50{,}000 & $128{\rightarrow}128$ \\
  & \method{} & 20{,}000 & $+$30{,}000 streaming & 50{,}000 & $128{\rightarrow}128$ \\
\bottomrule
\end{tabular}%
}
\end{table}
The FastWAM-Joint rows use ordinary shared-timestep flow matching, whereas the
StreamingWAM rows use the 80/20 mixture above.  All other methods retain the
training values reported in their original papers---including their reported
optimization budget, batch size, and learning-rate settings---rather than
being silently retrained with our recipe.  In particular, Fast-WAM follows
\citet{yuan2026fastwam}, Faster-WAM follows \citet{ma2026fasterwam}, and the
VLA baselines follow their cited original implementations.  We do not infer
an unreported hyperparameter.

\subsection{Evaluation and inference configuration}
\label{app:inference_config}

LIBERO-Plus results use the full 10{,}030-episode protocol
(\textsc{object} 2{,}518 / \textsc{spatial} 2{,}402 / \textsc{libero-10}
2{,}519 / \textsc{goal} 2{,}591).  Success is determined by the environment's
built-in criterion with no tolerance modification.  RoboTwin~2.0 results use
all 50 tasks with seed~0 under both clean and randomised initialisations,
following the official leaderboard protocol.  Real-robot demonstrations are
collected with a GELLO joint-level teleoperation device at 15\,Hz.  The training
set contains 590 trajectories across seven collection tasks: block stacking,
open drawer, close drawer, weigh-and-place, conveyor handover, car pursuit, and
whack-a-mole. Open and close drawer are collected separately but belong to the
same manipulation family. Evaluation reports the open-drawer variant and omits
the easier close-drawer variant, yielding the six evaluation tasks described
below. The number of trajectories varies by task, with some tasks containing
fewer than 100 demonstrations. Evaluation uses
50 trials per manipulation task and $50\!\times\!30$\,s sessions for
whack-a-mole.  Within $\pi_{0.5}$ and FastWAM-Joint, Async and RTC variants
share a checkpoint and differ only in execution strategy.

Introduced by \citet{lu2026faster}, TTFA measures the wall-clock time from an
episode's initial observation to its first executable action at the controller.
For \method{}, this once-per-episode quantity includes the warm start that
initializes the persistent action buffer. Normal closed-loop control does not
repeat that cost, so our primary latency metric is steady-state replanning time;
the initialization latency is reported separately. All measurements use the
same RTX~5090 GPU with CUDA-synchronous timing.

During simulation evaluation, \method{} uses synchronous control: the
environment waits for each policy call before executing the returned
$K$-action block.  Real-robot deployment instead uses asynchronous control.
For $K=4$, the next inference is launched only after the first three actions
of the current block have finished, so inference overlaps with execution of
the fourth action.

\begin{table*}[t]
\centering
\caption{Inference hyperparameters used in our evaluations.  $H$ is the
  predicted future-action horizon (the 16 clean prefix actions of
  StreamingWAM are excluded), and $K$ is the number of actions executed per
  replan.  $10\!\rightarrow\!1$ denotes a one-time 10-step warm start followed
  by one rolling SDP update per replan.  For real-robot non-streaming rows,
  $10/1$ means static/dynamic tasks.}
\label{tab:inference_config}
\setlength{\tabcolsep}{3pt}
\scriptsize
\begin{tabular*}{\textwidth}{@{\extracolsep{\fill}}l*{9}{c}@{}}
\toprule
& \multicolumn{3}{c}{LIBERO / LIBERO-Plus} & \multicolumn{3}{c}{RoboTwin~2.0} & \multicolumn{3}{c}{Real robot} \\
\cmidrule(lr){2-4}\cmidrule(lr){5-7}\cmidrule(lr){8-10}
Method & Steps & $H$ & $K$ & Steps & $H$ & $K$ & Steps & $H$ & $K$ \\
\midrule
$\pi_{0.5}$ Async & 10 & 10 & 5 & 10 & 50 & 50 & 10 & 16 & 4 \\
$\pi_{0.5}$ RTC & 10 & 10 & 5 & 10 & 50 & 50 & 10 & 16 & 4 \\
\fastwam & 10 & 32 & 10 & 10 & 32 & 24 & $10/1$ & 32 & 4 \\
\fasterwam & 10 & 32 & 4 & 10 & 32 & 4 & 10 & 32 & 4 \\
\fwjoint Async & 10 & 32 & 10 & 10 & 32 & 24 & $10/1$ & 32 & 4 \\
\fwjoint RTC & 10 & 32 & 4 & 10 & 32 & 4 & 10 & 32 & 4 \\
\textbf{\method{}} & $10\!\rightarrow\!1$ & 16 & 4 & $10\!\rightarrow\!1$ & 16 & 8 & $10\!\rightarrow\!1$ & 16 & 4 \\
\bottomrule
\end{tabular*}
\end{table*}

The $\pi_{0.5}$ simulation entries use the public implementation defaults:
LIBERO predicts 10 actions and executes five per query, while RoboTwin predicts
and executes 50; both use 10 flow-matching integration steps.  Async and RTC
therefore share the same simulation defaults and differ only in the real-robot
execution stack.

The RoboTwin Joint setting executes $K=24$ actions, whereas StreamingWAM uses
$K=8$; the latter value is an execution setting and must not be confused with
the training set $\mathcal{K}_{\mathrm{train}}$. The real-robot FastWAM-Joint
RTC row is the configuration reported in \cref{fig:real_results,tab:real_detail}.
The FastWAM-Joint LIBERO and \liberoplus result in \cref{tab:main} follows its
official configuration with $K=10$.
StreamingWAM's steady-state one is a rolling update of
a persistent buffer, not one-step denoising of an independently redrawn Joint
chunk.

\subsection{Baselines}
\label{app:baselines}

The LIBERO comparison includes 13 baselines spanning both VLA policies and
WAMs, grouped according to whether they use embodied data during pretraining.
Eight methods use embodied-data pretraining:
$\pi$0.5~\citep{black2025pi05},
GE-ACT~\citep{liao2025genieenvisioner},
VLA-JEPA~\citep{sun2026vlajepa},
X-VLA~\citep{zheng2025xvla},
OpenVLA-OFT~\citep{kim2025openvlaoft},
$\pi$0-Fast~\citep{pertsch2025fast},
$\pi$0~\citep{black2024pi0}, and
UniVLA~\citep{bu2025univla}.
The remaining five do not:
\fasterwam~\citep{ma2026fasterwam},
4D-WAM~\citep{yang20264dwam},
\fwjoint,
WorldVLA~\citep{cen2025worldvla},
\fastwam~\citep{yuan2026fastwam}.
The RoboTwin comparison additionally includes
AHA-WAM~\citep{cai2026ahawam}.
Among these, \fwjoint, the jointly trained variant of \fastwam, serves as our
closest direct baseline because it shares the same model architecture with
\method{} while omitting the three modules introduced in
\cref{sec:method:model}.

\section{Extended simulation results}
\label{app:extended}

\subsection{Inference latency}
\label{app:latency}

\Cref{tab:latency_detail} lists the steady-state per-replan latency for all
open-source WAM variants measured on the same RTX~5090 GPU with CUDA
synchronous timing.
Baseline architectures are reproduced from the published code with random
weights (network structure determines latency, not learned weights); our
model uses the exact checkpoint reported in the paper.

\begin{table}[h]
\centering
\caption{Inference latency comparison.  All measurements on the same RTX~5090
  with CUDA synchronous timing.  $n$: number of timed forward passes.}
\label{tab:latency_detail}
\setlength{\tabcolsep}{2.5pt}
\footnotesize
\begin{tabular*}{\linewidth}{@{\extracolsep{\fill}}lccrcr@{}}
\toprule
Method & $H$ & Steps & $n$ & Latency (ms) & \liberoplus (\%) \\
\midrule
\fwjoint        & 32 & 10 & 30  & 381.6\,$\pm$\,6.9 & 68.8 \\
\fastwam        & 32 & 10 & 30  & 182.6\,$\pm$\,3.2 & 51.5 \\
\fasterwam      & 32 & 10 & 30  & 139.4\,$\pm$\,1.7 & 75.0 \\
\textbf{Ours} (steady) & 16 & 1  & 300 & \textbf{72.9}\,$\pm$\,0.3 & \textbf{79.4} \\
\midrule
\textbf{Ours} (initial warm start) & 16 & 10 & 30 & 406.2\,$\pm$\,0.8 & --- \\
\midrule
\multicolumn{6}{@{}l}{\textit{Real-robot deployment (same RTX~5090)}} \\
\fwjoint        & 32 & 10 & 30  & 368.6\,$\pm$\,2.4 & --- \\
\fwjoint (timing only) & 32 & 1 & 30 & 46.8\,$\pm$\,0.3 & --- \\
\fastwam        & 32 & 10 & 30  & 182.6\,$\pm$\,3.2 & --- \\
$\pi_{0.5}$     & 16 & 10 & 30  & 61.5\,$\pm$\,1.2  & --- \\
\textbf{Ours} (steady)     & 16 & 1  & 300 & \textbf{57.3}\,$\pm$\,0.6 & --- \\
\textbf{Ours} (initial warm start) & 16 & 10 & 30  & 436.8\,$\pm$\,1.6 & --- \\
\bottomrule
\end{tabular*}
\end{table}
The 1-step FastWAM-Joint timing row uses the same checkpoint as the 10-step row.
One StreamingWAM forward is not cheaper than one Joint forward (57.3 versus
46.8\,ms).  Its steady-state advantage comes from the persistent buffer, which
reduces each replan from ten model evaluations to one.  Because the reported
deployment configurations also differ in $H$, these speedups are system-level
rather than a controlled single-variable comparison.  In the real-robot task
results, Fast-WAM and FastWAM-Joint use 10 steps on static tasks and one step in
the dynamic-task main setting, as summarized in \cref{tab:inference_config}.
The one-step dynamic setting is necessary because preliminary 10-step
deployments yield near-zero success at the required reaction time scale.

\subsection{Action continuity}
\label{app:smoothness}

\Cref{tab:smoothness} reports the action-continuity metrics underlying
\cref{fig:smoothness}.
All three models are evaluated with 1-step inference and $\Kexec{=}4$ on a
500-episode balanced subset of \liberoplus.
Translation and rotation jumps are the $\ell_2$ norm of the end-effector
command change at replanning boundaries; the boundary-to-interior ratio
measures the ratio of the jump at step~0 to the average jump at non-boundary
steps; global translation jerk is the mean $\ell_2$ norm of the finite-
differenced acceleration of end-effector translation commands.

\begin{table}[h]
\centering
\caption{Action-continuity metrics on \liberoplus (500-episode subset,
  1-step inference, $\Kexec{=}4$).  $\downarrow$: lower is better.}
\label{tab:smoothness}
\setlength{\tabcolsep}{3pt}
\footnotesize
\begin{tabular*}{\linewidth}{@{\extracolsep{\fill}}lccccr@{}}
\toprule
Method & SR (\%) & Trans.\ jump $\downarrow$ & Rot.\ jump $\downarrow$ & Bnd./int.\ ratio $\downarrow$ & Jerk $\downarrow$ \\
\midrule
\fastwam        & 60.2 & 0.1350 & 0.02903 & 1.831 & 5.775 \\
\fwjoint        & 68.8 & 0.1379 & 0.02681 & 1.840 & 5.710 \\
\textbf{Ours}   & \textbf{79.0} & \textbf{0.0684} & \textbf{0.00879} & \textbf{1.105} & \textbf{2.341} \\
\midrule
\multicolumn{2}{l}{Ours vs.\ \fwjoint} & $-$50.4\% & $-$67.2\% & --- & $-$59.0\% \\
\bottomrule
\end{tabular*}
\end{table}

\subsection{Per-suite breakdown on \liberoplus}
\label{app:persuite}

\Cref{tab:persuite} breaks down the success rate by suite and
perturbation type for \fwjoint (Joint) and our model.
Our model improves on every suite, with the largest gains on
\textsc{object} (+16.4\,pp) and \textsc{libero-10} (+13.8\,pp).
Across perturbation types, the gap is widest under camera-viewpoint
(+18.5\,pp pooled) and robot-initial-state (+18.2\,pp) perturbations,
where \fwjoint is most vulnerable.

\begin{table}[h]
\centering
\caption{Per-suite per-perturbation success rate (\%) on \liberoplus
  (10{,}030 episodes). Ours uses $\Kexec{=}4$;
  Joint denotes the official-configuration \fwjoint baseline from
  \cref{tab:main}.
  BG: background textures; Robot: robot initial states;
  Cam: camera viewpoints; Lang: language instructions;
  Noise: sensor noise; Obj: objects layout; Light: light conditions.}
\label{tab:persuite}
\setlength{\tabcolsep}{2pt}
\scriptsize
\begin{tabular*}{\linewidth}{@{\extracolsep{\fill}}llrrrrrrrr@{}}
\toprule
Suite & & BG & Robot & Cam & Lang & Noise & Obj & Light & All \\
\midrule
\multirow{2}{*}{\textsc{object}}
  & Joint & 64.1 & 57.8 & 49.2 & 97.7 & 68.0 & 77.4 & 97.3 & 72.2 \\
  & \textbf{Ours} & 83.9 & 82.2 & 83.6 & 97.7 & 92.2 & 83.9 & 98.0 & \textbf{88.6} \\
\cmidrule(lr){1-10}
\multirow{2}{*}{\textsc{spatial}}
  & Joint & 74.0 & 66.3 & 55.6 & 93.3 & 72.1 & 90.4 & 99.0 & 78.5 \\
  & \textbf{Ours} & 73.6 & 83.7 & 64.4 & 97.4 & 87.5 & 96.4 & 97.9 & \textbf{86.1} \\
\cmidrule(lr){1-10}
\multirow{2}{*}{\textsc{libero-10}}
  & Joint & 32.2 & 72.3 & 30.8 & 96.6 & 64.1 & 87.8 & 84.7 & 66.3 \\
  & \textbf{Ours} & 53.6 & 88.3 & 54.7 & 100.0 & 76.6 & 92.9 & 98.5 & \textbf{80.1} \\
\cmidrule(lr){1-10}
\multirow{2}{*}{\textsc{goal}}
  & Joint & 54.4 & 48.2 & 35.8 & 84.9 & 48.8 & 61.6 & 85.3 & 59.0 \\
  & \textbf{Ours} & 50.9 & 62.8 & 42.6 & 74.4 & 57.5 & 64.9 & 96.8 & \textbf{63.4} \\
\midrule
\multirow{2}{*}{\textbf{Pooled}}
  & Joint & 55.4 & 60.8 & 42.5 & 92.9 & 63.3 & 78.4 & 91.8 & 68.8 \\
  & \textbf{Ours} & 64.7 & 79.0 & 61.0 & 92.0 & 78.6 & 83.6 & 97.8 & \textbf{79.4} \\
\bottomrule
\end{tabular*}
\end{table}

\section{Per-task results on RoboTwin~2.0}
\label{app:robotwin_pertask}

\Cref{tab:robotwin_pertask} reports the per-task success rates of
\method{} under the clean and randomized object-pose settings.  Averaging
the 50 tasks gives 84.2\% and 5.1\%, respectively, matching the aggregate
results in \cref{tab:robotwin}.

\begin{center}
\centering
\captionof{table}{Per-task success rate (\%) of \method{} on all 50 RoboTwin~2.0
  tasks.  The table is arranged in two blocks for compactness.}
\label{tab:robotwin_pertask}
\setlength{\tabcolsep}{3pt}
\scriptsize
\begin{tabular*}{\textwidth}{@{\extracolsep{\fill}}lrrlrr@{}}
\toprule
Task & Clean & Randomized & Task & Clean & Randomized \\
\midrule
\texttt{adjust\_bottle}              & 100.0 &  2.0 & \texttt{place\_can\_basket}          & 60.0 &  6.0 \\
\texttt{beat\_block\_hammer}        &  93.0 &  1.0 & \texttt{place\_cans\_plasticbox}     & 96.0 &  6.0 \\
\texttt{blocks\_ranking\_rgb}       &  99.0 &  2.0 & \texttt{place\_container\_plate}     & 99.0 &  4.0 \\
\texttt{blocks\_ranking\_size}      &  45.0 &  0.0 & \texttt{place\_dual\_shoes}          & 86.0 &  2.0 \\
\texttt{click\_alarmclock}           & 100.0 & 40.0 & \texttt{place\_empty\_cup}           & 91.0 &  5.0 \\
\texttt{click\_bell}                 & 100.0 & 17.0 & \texttt{place\_fan}                   & 80.0 &  0.0 \\
\texttt{dump\_bin\_bigbin}          &  93.0 &  1.0 & \texttt{place\_mouse\_pad}           & 54.0 &  0.0 \\
\texttt{grab\_roller}                &  98.0 &  5.0 & \texttt{place\_object\_basket}       & 87.0 &  3.0 \\
\texttt{handover\_block}             &  98.0 &  0.0 & \texttt{place\_object\_scale}        & 81.0 &  0.0 \\
\texttt{handover\_mic}               &  99.0 &  3.0 & \texttt{place\_object\_stand}        & 94.0 &  7.0 \\
\texttt{hanging\_mug}                &  18.0 &  0.0 & \texttt{place\_phone\_stand}         & 83.0 &  0.0 \\
\texttt{lift\_pot}                   &  94.0 &  4.0 & \texttt{place\_shoe}                  & 94.0 &  5.0 \\
\texttt{move\_can\_pot}             &  85.0 &  0.0 & \texttt{press\_stapler}               & 84.0 &  3.0 \\
\texttt{move\_pillbottle\_pad}      &  95.0 &  1.0 & \texttt{put\_bottles\_dustbin}       & 91.0 &  0.0 \\
\texttt{move\_playingcard\_away}    &  96.0 &  0.0 & \texttt{put\_object\_cabinet}        & 51.0 &  1.0 \\
\texttt{move\_stapler\_pad}         &  42.0 &  0.0 & \texttt{rotate\_qrcode}               & 86.0 &  0.0 \\
\texttt{open\_laptop}                &  85.0 &  0.0 & \texttt{scan\_object}                 & 87.0 &  4.0 \\
\texttt{open\_microwave}             &  64.0 &  5.0 & \texttt{shake\_bottle}                &100.0 & 28.0 \\
\texttt{pick\_diverse\_bottles}     &  75.0 &  9.0 & \texttt{shake\_bottle\_horizontally} &100.0 & 22.0 \\
\texttt{pick\_dual\_bottles}        &  85.0 & 11.0 & \texttt{stack\_blocks\_three}        & 95.0 &  0.0 \\
\texttt{place\_a2b\_left}           &  95.0 &  3.0 & \texttt{stack\_blocks\_two}          & 99.0 & 12.0 \\
\texttt{place\_a2b\_right}          &  90.0 &  1.0 & \texttt{stack\_bowls\_three}         & 83.0 &  9.0 \\
\texttt{place\_bread\_basket}       &  89.0 & 13.0 & \texttt{stack\_bowls\_two}           & 90.0 &  4.0 \\
\texttt{place\_bread\_skillet}      &  90.0 &  7.0 & \texttt{stamp\_seal}                  & 65.0 &  0.0 \\
\texttt{place\_burger\_fries}       &  92.0 &  8.0 & \texttt{turn\_switch}                 & 54.0 &  1.0 \\
\midrule
\textbf{Mean}                         & \textbf{84.2} & \textbf{5.1}
  & \multicolumn{3}{c}{} \\
\bottomrule
\end{tabular*}
\end{center}

\section{Additional ablation results}
\label{app:additional_ablation}

\subsection{Component contributions at \texorpdfstring{$\Kexec=4$}{K=4}}
\label{app:component_ablation}

\begin{figure}[h]
\centering
\includegraphics[width=0.48\linewidth]{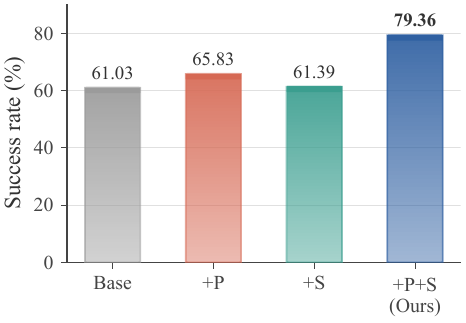}
\caption{Component contributions on \liberoplus at $\Kexec=4$. The clean
  action prefix with dense visual history~(P) and the streaming staircase~(S)
  have complementary effects and combine to achieve the strongest result.}
\label{fig:app_ablation_bar}
\end{figure}

\subsection{Warm-start ablation}
\label{app:warmstart}

\Cref{sec:exp:ablation} reports the aggregate effect of warm-start
length.  \Cref{tab:warmstart_full} gives the per-setting breakdown on
the full 10{,}030-episode evaluation: removing the warm start drops
success by 5.2\,pp, while a single step already recovers the gap.

\begin{table}[h]
\centering
\caption{Warm-start ablation on \liberoplus at $\Kexec{=}4$.}
\label{tab:warmstart_full}
\setlength{\tabcolsep}{6pt}
\footnotesize
\begin{tabular*}{\linewidth}{@{\extracolsep{\fill}}rccc@{}}
\toprule
Warm-start steps & Successes & Total & Success (\%) \\
\midrule
0  & 7{,}443 & 10{,}030 & 74.2 \\
1  & 7{,}934 & 10{,}030 & 79.1 \\
4  & 7{,}919 & 10{,}030 & 79.0 \\
8  & 7{,}927 & 10{,}030 & 79.0 \\
10 & 7{,}960 & 10{,}030 & 79.4 \\
\bottomrule
\end{tabular*}
\end{table}

\section{Real-robot experiment details}
\label{app:real_tasks}

\subsection{Task descriptions}
\label{app:real_task_descriptions}

We evaluate six manipulation tasks, grouped by whether the target object moves
during execution.  \emph{Static tasks}: \textbf{block stacking}---stack five
layers of blocks; \textbf{open drawer}---grasp the handle and pull the drawer
open; and \textbf{weigh-and-place}---place an apple on a kitchen
scale, read the weight, then transfer it to a bowl.  \emph{Dynamic tasks}:
\textbf{conveyor handover}---pick a moving apple off a conveyor belt and place
it into a bowl; \textbf{car pursuit}---intercept an apple riding a curved-path
toy car and place it into a bowl; and \textbf{whack-a-mole}---press the
randomly lit button on a response-time toy.  The dynamic tasks require the
policy to continuously adapt its actions to moving objects and therefore
directly evaluate reactive control in time-varying environments.

\subsection{Observation preprocessing}
\label{app:real_obs}

\begin{figure}[h]
\centering
\includegraphics[width=0.3\linewidth]{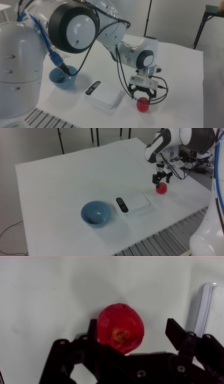}
\caption{Real-robot model input: three camera views (exterior~1,
  exterior~2, wrist) vertically concatenated into a
  $384{\times}224$ frame ($H{\times}W$).}
\label{fig:real_obs}
\end{figure}

\Cref{fig:real_obs} shows a representative model input frame from the
real robot.  Each of the three camera views (two exterior RealSense D435i
and one wrist-mounted ZED Mini) is resized to $128{\times}224$ and
vertically stacked into a $384{\times}224$ tensor (all dimensions are
$H{\times}W$).
In simulation, LIBERO provides two views horizontally concatenated into a
$224{\times}448$ tensor.

\subsection{Action continuity on the real robot}
\label{app:real_smoothness}

\Cref{fig:latency_smoothness}(d) compares aligned real-robot trajectory
segments from FastWAM-Joint under asynchronous and RTC execution with
StreamingWAM.  The upper trace reports joint displacement from the
initial posture, while the lower trace reports the supplied discrete
acceleration proxy.  StreamingWAM remains lower than both Joint baselines
for most samples; we do not interpret the proxy as physical acceleration
or jerk.

\subsection{Qualitative moving-target anticipation}
\label{app:real_motion_anticipation}

\begin{figure}[H]
  \centering
  \includegraphics[width=0.98\linewidth]{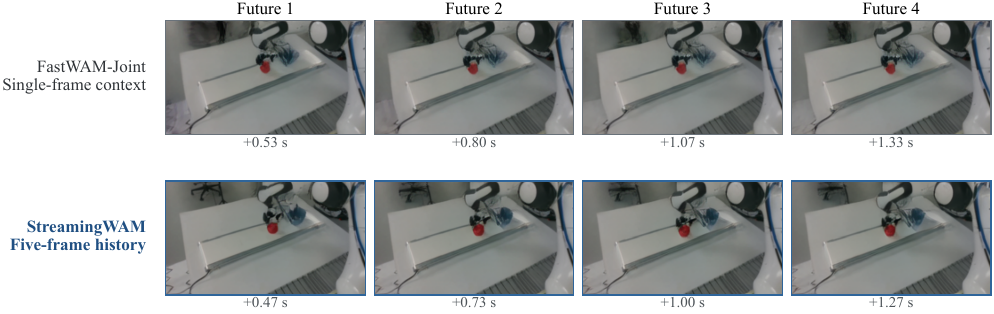}
  \caption{Qualitative moving-target prediction before contact. Four future
  frames are replayed offline from recorded conveyor-task inputs and action
  chunks. FastWAM-Joint uses single-frame
  context; \method{} uses five-frame dense history. In external camera~3, the
  predicted/recorded apple displacement is 1.38/6.93 pixels for FastWAM-Joint
  and 9.66/5.93 pixels for \method{}; the cosine between predicted and recorded
  displacement directions is 0.497 and 0.995, respectively. The examples are
  stage-matched independent trials, so they support a qualitative anticipation
  interpretation but do not isolate history as the sole cause.}
  \label{fig:real_motion_anticipation}
\end{figure}

To further examine why the largest real-robot gains occur on dynamic tasks,
\cref{fig:real_motion_anticipation} visualizes future video predictions before
contact in stage-matched conveyor trials.  FastWAM-Joint largely freezes the
apple, whereas \method{} predicts displacement along the conveyor direction.
The examples come from independent trials and the two models differ in more
than their visual context, so this evidence supports an anticipation
interpretation but does not isolate history as the sole cause.

\subsection{Per-task results}
\label{app:real_pertask}

\Cref{tab:real_detail} reports per-task results for all six real-robot
tasks.  The five manipulation tasks are evaluated over 50 trials each;
whack-a-mole reports valid activations of the currently lit button across
50$\times$30\,s sessions. An incorrect or repeated press is a failure; reaching
the device's failure limit triggers a reset that consumes part of the session.

\begin{table}[H]
\centering
\caption{Real-robot per-task results. Number of successful trials for the five
  manipulation tasks (50 trials each); valid lit-button activations for
  whack-a-mole (50$\times$30\,s sessions).}
\label{tab:real_detail}
\setlength{\tabcolsep}{2.2pt}
\scriptsize
\begin{tabular*}{\linewidth}{@{\extracolsep{\fill}}lrrrrrr@{}}
\toprule
& \multicolumn{2}{c}{$\pi_{0.5}$}
& \multicolumn{2}{c}{FastWAM-Joint}
& Fast-WAM & \textbf{StreamingWAM} \\
\cmidrule(lr){2-3}\cmidrule(lr){4-5}
Task & Async & RTC & Async & RTC & Async & \textbf{Async} \\
\midrule
\multicolumn{7}{@{}l}{\textit{Static tasks}} \\
Block stacking  & 29 & \textbf{31} & 16 & 18 & 13 & 28 \\
Open drawer     & 45 & \textbf{46} & 38 & 40 & 36 & \textbf{46} \\
Weigh-and-place & 23 & 25 & 18 & 19 & 15 & \textbf{42} \\
\addlinespace[1pt]
\multicolumn{7}{@{}l}{\textit{Dynamic tasks}} \\
Conveyor handover & 15 & 9 & 9 & 5 & 6 & \textbf{32} \\
Car pursuit       & 8 & 4 & 4 & 2 & 4 & \textbf{20} \\
\midrule
\textit{Five-task total} & 120 & 115 & 85 & 84 & 74 & \textbf{168} \\
Whack-a-mole (valid activations) & 391 & 408 & 369 & 357 & 330 & \textbf{424} \\
\bottomrule
\end{tabular*}
\end{table}

\subsection{Task execution keyframes}

\begin{figure}[H]
\centering
\captionsetup[subfigure]{font=small,skip=1pt}
\begin{subfigure}{\textwidth}
\centering
\includegraphics[width=\textwidth]{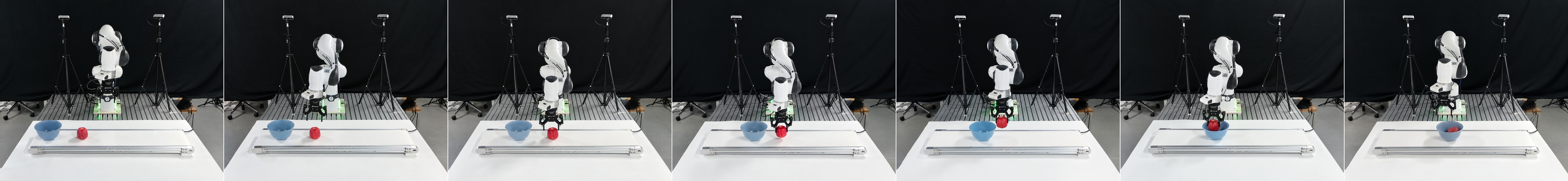}
\caption{Conveyor-belt handover.}\label{fig:real_conveyor}
\end{subfigure}
\vspace{-2pt}
\begin{subfigure}{\textwidth}
\centering
\includegraphics[width=\textwidth]{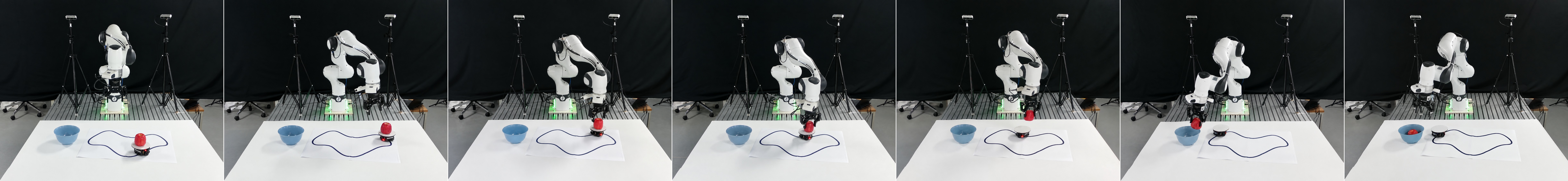}
\caption{Car pursuit.}\label{fig:real_car}
\end{subfigure}
\vspace{-2pt}
\begin{subfigure}{\textwidth}
\centering
\includegraphics[width=\textwidth]{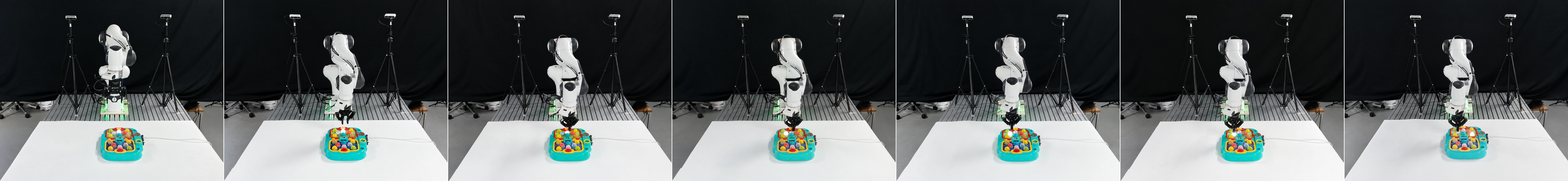}
\caption{Whack-a-mole.}\label{fig:real_whack}
\end{subfigure}
\vspace{-2pt}
\begin{subfigure}{\textwidth}
\centering
\includegraphics[width=\textwidth]{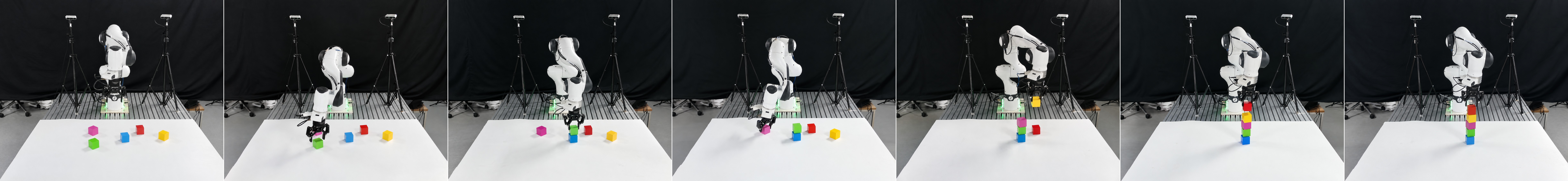}
\caption{Block stacking.}\label{fig:real_blocks}
\end{subfigure}
\vspace{-2pt}
\begin{subfigure}{\textwidth}
\centering
\includegraphics[width=\textwidth]{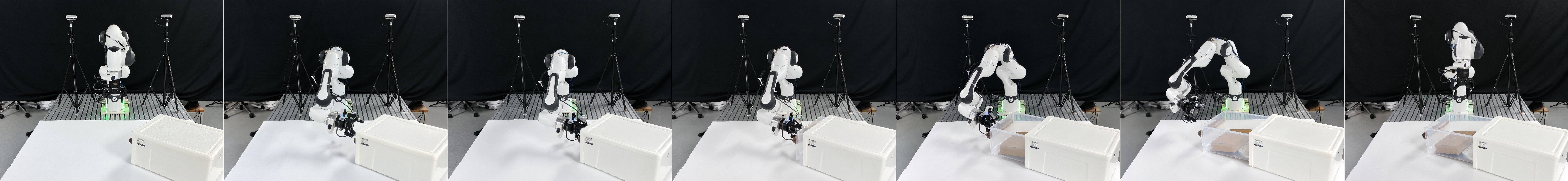}
\caption{Open drawer.}\label{fig:real_drawer}
\end{subfigure}
\vspace{-2pt}
\begin{subfigure}{\textwidth}
\centering
\includegraphics[width=\textwidth]{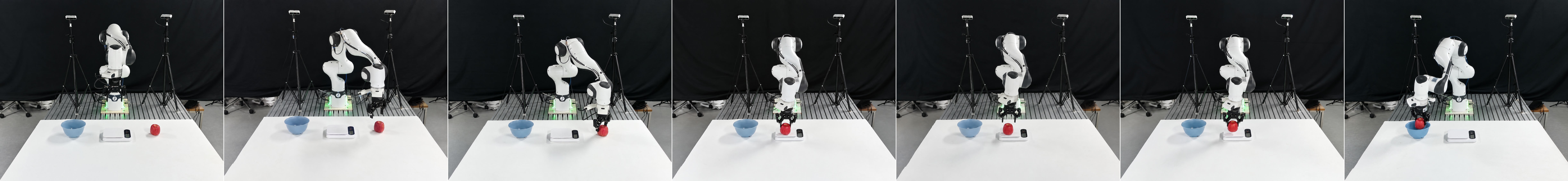}
\caption{Weigh-and-place.}\label{fig:real_weigh}
\end{subfigure}
\caption{Real-robot task execution under streaming single-step replanning.
Each row shows seven keyframes from one successful episode.}
\label{fig:real_keyframes}
\end{figure}

\Cref{fig:real_conveyor,fig:real_car,fig:real_whack,fig:real_blocks,fig:real_drawer,fig:real_weigh}
show successful rollouts of all six real-robot tasks under streaming
single-step replanning.

\end{document}